\documentclass[11pt]{article}
\usepackage[letterpaper,margin=1in]{geometry}
\usepackage[T1]{fontenc}
\usepackage{graphicx}
\usepackage{amsmath,amssymb,amsthm}
\usepackage{array}
\usepackage{float}
\usepackage{url}
\usepackage[hidelinks]{hyperref}

\newtheorem{proposition}{Proposition}
\title{A Contract-Centered Architecture for Scalable and Manageable Agentic Runtimes}
\author{%
Yaxiao Liu\\
PwC China AI Center\\
\texttt{yaxiao.y.liu@cn.pwc.com}\\
\texttt{rootliu@gmail.com}
\and
Pengbo Liu\\
PwC China AI Center\\
\texttt{liupengbo@mails.neu.edu.cn}
\and
Yiwen Liu\\
PwC China AI Center\\
\texttt{202383049@uibe.edu.cn}
\and
Yihua Guan\\
PwC China AI Center\\
\texttt{202311260036@mail.bnu.edu.cn}
\and
Zhenghe Hou\\
PwC China AI Center\\
\texttt{zh3773@nyu.edu}
\and
Jiaxing Song\\
Tsinghua University\\
\texttt{jxsong@tsinghua.edu.cn}
}
\date{}

\usepackage{longtable}
\newcolumntype{P}[1]{>{\raggedright\arraybackslash}p{\dimexpr #1\linewidth-2\tabcolsep\relax}}
\begin{document}
\maketitle

\begin{abstract}

Enterprise agentic systems change along two axes: the capabilities admitted for a task and the capacity available to run them. We assign business behavior to Skills, logical admission and verification to the Harness, and physical execution to the Scaffold. An external data substrate governs source identity, access and evidence. Our central hypothesis asks whether compatible capacity changes preserve semantic outcomes and whether capability changes preserve capacity response within declared margins and enforcement budgets.

We specify how to decide whether a worker may replace another on an admitted path. The contract pins the execution profile, authority and placement constraints. At dispatch, the runtime checks the observed worker and records what actually executed. After an ambiguous write timeout, recovery depends on the effect receipt and the scope and lifetime of deduplication. A reference fixture agrees with its declared oracle on all 256 synthetic binding combinations and passes a compatible-replica control and ten write-recovery checks.

On the data boundary, executable 5W1H+Which predicates bind source versions, authorization and evidence. A separate single-process model agrees with its oracle on 1,024 synthetic fault combinations and passes five lifecycle checks. Both artifacts establish finite conformance under trusted inputs. A held-out data study and a cluster-period crossover specify how to measure retrieval quality, semantic non-inferiority, capacity response and enforcement cost. The separability hypothesis remains empirically untested.

\end{abstract}

\section{Introduction}
\label{sec:introduction}

Enterprise AI is usually assembled along existing organizational boundaries. A business unit owns a procurement, customer-service, research, or finance outcome. Application and automation developers encode the workflow. AI platform teams change models and agent patterns. Testing and governance teams define release evidence. Cloud, server-farm, site-reliability engineering (SRE), security, and operations teams supply and control execution. The CIO and domain data stewards govern meaning, access, lineage, and lifecycle across enterprise systems. These groups change different objects at different rates, yet an agent run crosses all of them.

Use-case evaluation alone does not close this deployment gap. A benchmark can establish task success, requirement completion, quality, cycle time, or human correction for a particular Skill. It cannot by itself establish that the capability was admitted under the intended policy, composed safely, bound to compatible capacity, isolated from other tenants, replayable, portable, or affordable at enterprise scale. Conversely, a runtime can meet throughput and availability targets while the Skill produces no business value. Business/use-case evidence and system/runtime evidence are therefore separate and neither substitutes for the other.

An agentic runtime must answer two different change requests. A product owner wants to activate a new behavior without redesigning the execution fleet. A platform owner wants to add workers, isolation zones, accelerators, or regional capacity without changing what an admitted behavior means. Current architecture descriptions often name both concerns but evaluate them separately: capability work is judged by task success, while infrastructure work is judged by throughput and latency. Those one-axis evaluations cannot reveal whether the axes recouple through shared state, resource-sensitive inputs, scheduler feedback, undeclared effects, or bypasses around policy enforcement.

The distinction matters because a positive capacity effect is desirable. Adding compatible Scaffold resources should normally improve throughput or queueing delay. The scientific question is not whether the Scaffold has a zero main effect. It is whether the response to a capability intervention materially changes across Scaffold configurations, whether semantic outcomes remain non-inferior, and whether the controls required to establish that result stay within an acceptable enforcement budget.

This paper therefore asks one research question:

\textit{Can a runtime activate independently deployable capabilities without materially changing capacity response, while scaling capacity without materially changing capability semantics, at an acceptable enforcement cost?}

We call the bounded claim \textit{cost-aware capability-capacity separability}. It is neither universal modularity nor proof that a named implementation scales. It is an empirical decision about a specified runtime, workload, policy, data snapshot, and compatible capacity range, and its verdict can be supported within $\Omega$, falsified within $\Omega$, reported as a conditional-engineering result, or inconclusive.

Alongside that scientific question the paper advances a narrower organizational thesis of its own: for enterprise AI, architecture can serve as a shared organizational contract that makes communication, ownership, change cadence, and evidence obligations explicit across business and technology groups. This is not an external quotation. It is the organizing interpretation used here to connect business capability, runtime governance, execution capacity, and enterprise data without requiring one mandatory organization chart.

The architecture is organized around a compact model:

\begin{equation}
\mathcal{A}=\langle\mathcal{S},\mathcal{H},\mathcal{X}\rangle,
\end{equation}

where $\mathcal{S}$ is a versioned set of Skills, $\mathcal{H}$ is the Harness that compiles and governs runs, and $\mathcal{X}$ is a pool of Scaffold instances. Skill owns bounded business semantics. Harness owns runtime admission, composition, governance, and evidence capture. Scaffold owns the physical execution and control boundary and its NFRs. A stack-external data substrate, independently governed by the CIO or equivalent data authority, supplies semantic joins, governed access, provenance, routing facts, intermediate representations, and integrated business and runtime telemetry.

\begin{figure}[t]
\centering
\includegraphics[width=\linewidth]{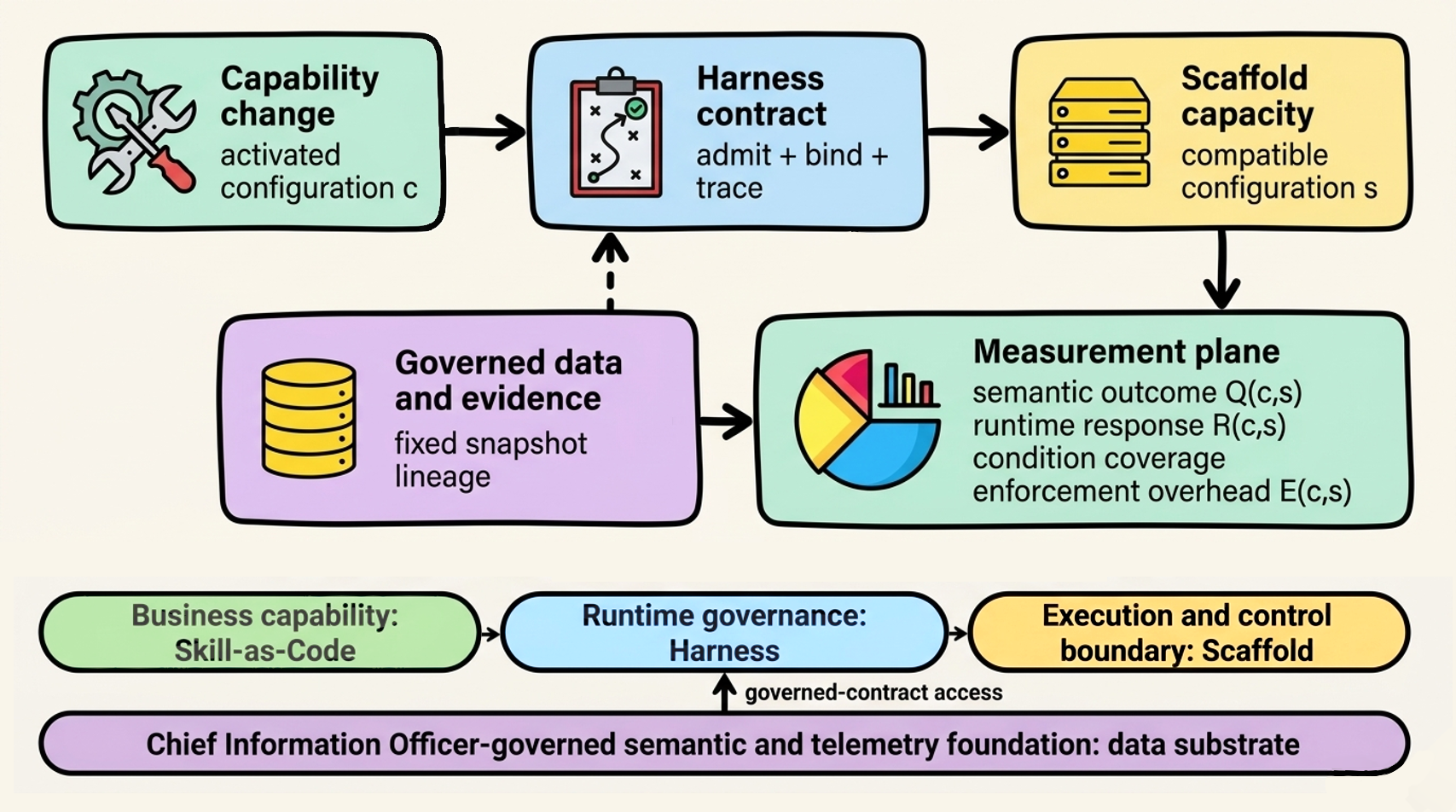}

\caption{Shown: business capability enters as a versioned Skill contract, the Harness admits and binds an activated path, the Scaffold supplies the execution and control boundary, and the CIO-governed semantic and telemetry substrate remains stack-external; any derived index or summary in that substrate is discovery metadata rather than source authority. Why it matters: the diagram separates the two interventions whose independence P1 tests and identifies the boundary at which recoupling and enforcement cost must be observed. Class: architecture.}
\label{fig:dual-scaling}
\end{figure}

Figure~\ref{fig:dual-scaling} establishes both the responsibility handoffs and the scientific distinction. Capability growth means activated independently deployable behavior on admitted paths; it does not mean the number of entries in a registry. Capacity growth means a change to compatible physical execution resources under fixed logical policy and fixed data snapshots. The Harness is the binding boundary that makes both interventions inspectable.

\subsection{Contributions}

This paper makes three contributions.

\begin{enumerate}
\item We define a contract-bounded dual-axis responsibility architecture: reusable business capability, runtime governance, physical execution and NFR ownership, and enterprise data governance remain distinguishable, while the Harness resolves a typed contract $C=\langle I,O,G,A,B,V\rangle$ and specifies which worker can execute each admitted node. Dispatch-time binding and write-recovery controls make this boundary executable (Section~\ref{sec:scaffold-binding}).
\item We specify a source-preserving address/index/operator data substrate. Canonical sources and versioned provenance remain authoritative; derived summaries are auditable metadata rather than replacements; and versioned Theme/IR policies state task-family output, access, evidence, fallback, and compatibility requirements. An executable data-use ticket adds execution-time version and policy revalidation, typed rejection, and evidence binding, with a finite conformance artifact. A technical appendix formalizes the boundary for deterministic query-agnostic write-time representations through a decoder-independent Fano bound, a partition-lattice sufficiency criterion, and a linear-logarithmic write/read separation.
\item We turn six familiar design conditions into measured obligations and specify a cluster-period randomized crossover that estimates the capability-by-capacity interaction, semantic non-inferiority, and enforcement overhead without treating an expected Scaffold main effect as evidence of separability. The resulting evidence protocol makes the architecture rejectable rather than presenting its mechanisms as observed results.
\end{enumerate}

The two artifacts exercise the data-use and physical binding decisions on finite inputs. The crossover estimates the consequences of those decisions under changes in load and capability. Table~\ref{tab:propositions} and Section~\ref{sec:limitations} record the evidence status of every claim; the body states each mechanism once rather than repeating that classification alongside it.

\section{Origins: Organization, Boundaries, and Change Cadence}
\label{sec:origins}

\subsection{Organization, Communication, and Architecture}

Conway's 1968 observation is that organizations tend to produce system structures reflecting their communication structures \cite{conway1968committees}. Enterprise AI makes that observation operationally relevant because one run crosses business ownership, application development, AI runtime governance, testing, infrastructure, security, operations, and data stewardship. A hidden or ambiguous handoff in the organization can become a hidden or ambiguous dependency in the system.

This paper draws a narrower conclusion of its own. Architecture can be used as a shared organizational contract that states who owns an object, what that object promises, what it does not own, how quickly it may change, and what evidence must cross each boundary. The claim is not that every enterprise needs four teams or four processes. It is that business semantics, runtime governance, physical execution, and enterprise data require distinguishable decision rights even when one team implements several of them.

\subsection{Ownership and Change Cadence}

The four responsibility objects differ most visibly in ownership and expected rate of change. Table~\ref{tab:ownership-cadence} is a responsibility model rather than a mandatory organization chart.

\begin{table}[t]
\centering
\caption{Enterprise ownership, typical change cadence, and stable contract. One team may implement multiple objects, but their decision rights, stable contracts, and evidence responsibilities remain distinguishable.}
\label{tab:ownership-cadence}
\begin{tabular}{P{0.16}P{0.28}P{0.18}P{0.38}}
\hline
Object & Primary enterprise owner & Typical change cadence & Stable contract \\
\hline
Skill & Business product owner with AI/automation developers & Hours to weeks & Goal, typed I/O, effects, tests, evidence, version \\
Harness & AI platform/runtime and governance teams & Days to months & Admission, activated path, effects, binding constraints, postconditions, trace \\
Scaffold & Enterprise platform, cloud/server-farm, SRE, security, and operations & Weeks to years & Resource, isolation, locality, identity, attestation, execution \\
Data substrate & CIO/data authority with domain data stewards & Weeks to years & Semantic access, provenance, routing, lifecycle, intermediate representation \\
\hline
\end{tabular}
\end{table}

A business workflow or Skill may change within hours, while an isolation boundary, regional topology, or enterprise semantic standard may remain for years. Directly embedding slower-changing infrastructure and data assumptions in every Skill makes both sides expensive to change. The Harness contract instead translates an activated business capability into explicit runtime requirements and binds only accepted work to compatible capacity and governed data access.

\subsection{Runtime, Capacity, and External Data Boundaries}

The control-plane/data-plane distinction motivates, but does not prove, the proposed separation. Skill registration, release state, contract compilation, policy versions, and placement rules are control decisions. Accepted payload processing and effect execution are data-plane work. Selective activation keeps the full registry and governance state out of each request, while a resolved contract carries only the path, constraints, identities, and evidence obligations needed for that run.

Enterprise data has a different boundary again. The data substrate is not a fourth runtime layer and not a database hidden inside the Scaffold. It is stack-external, independently CIO-governed semantic and telemetry infrastructure with a typically slower change cadence than Skills. Skills consume its contracts; the Harness mediates access and evidence; the Scaffold enforces locality and network boundaries. Governed online access paths and asynchronous background indexing or maintenance loops can evolve behind that contract without requiring source-specific context assembly in every Skill.

\section{Foundations and Novelty Boundary}
\label{sec:related}

\paragraph{Classical foundations.}
The six obligations of Section~\ref{sec:obligations} are established systems ideas, and stating their lineage constrains the novelty claim. A reference monitor must be tamper resistant, always invoked, and small enough to analyze; complete mediation requires checking every access rather than only the first one \cite{anderson1972reference,saltzer1975protection}. Effect non-interference is a form of the information-flow non-interference long studied in security semantics \cite{goguen1982noninterference}; capability systems and least authority explain why an admitted path should carry narrow, unforgeable authority instead of ambient privilege \cite{miller2003capability}. Information hiding separates decisions likely to change behind stable interfaces \cite{parnas1972criteria}, Hydra made policy-mechanism separation an explicit operating-system principle \cite{levin1975hydra}, and control/data-plane separation distinguishes decisions from the execution of accepted work \cite{kreutz2015sdn}. Autonomic-computing work supplies the Monitor, Analyze, Plan, Execute over a shared Knowledge base (MAPE-K) control-loop vocabulary \cite{kephart2003autonomic,ibm2006autonomic}, and ISO/IEC 25010 supplies a product-quality vocabulary, not margins for this runtime \cite{iso25010}. Queueing and datacenter-interference results motivate tail-latency and saturation metrics \cite{kleinrock1975queueing,mars2011bubbleup,dean2013tail}. The paper adapts these ideas to agentic runtimes; it does not re-derive them.

\paragraph{The nearest thesis: scaling the harness.}
The closest published statement of this paper's premise argues that the next bottleneck in agentic AI is system scaling rather than model scaling, and names the object of that shift explicitly: treating the structured execution layer around a foundation model as a first-class object of design, evaluation, and optimization \cite{gu2026systemscaling}. That is the same object we call the Harness, and the agreement is worth stating plainly rather than minimizing, because it means the identification of the harness as the locus of scalability is not this paper's contribution. What remains ours is narrower and specific: that work argues for the shift and surveys its design space, whereas we commit to one falsifiable consequence of it --- that capability and compatible capacity are separable within a declared operating region at a bounded enforcement cost --- and specify the randomized design, margins, obligations, and four-state verdict that could refute it. A programmatic argument that the harness deserves first-class treatment and a preregistered protocol that can reject a specific separability claim about it are different kinds of contribution; conflating them would overstate ours.

\paragraph{Runtime governance and composition.}
The five-plane reference architecture organizes production-agent governance into distinct concerns and supplies contemporary support for complete mediation, explicit invariants, and structured evidence \cite{tallam2026fiveplane}. That work motivates enforceable runtime governance, but it does not establish a separability claim. Composition research shows why the activated path must be the authorization object: components benign in isolation can become harmful when composed \cite{xie2026composition}. Our Harness therefore evaluates the path together with its data, effect, identity, and resource bindings rather than treating a set of locally approved Skills as sufficient.

\paragraph{The closest existing implementation.}
Falsifiable release-gate work is the nearest published counterpart to the mechanism this paper proposes, and stating its reach precisely is what bounds our remaining contribution \cite{soni2026gates}. It predeclares a machine-checkable acceptance suite for every new capability, holds a fixed set of standing invariants across six consecutive releases of an open runtime, and exhaustively checks the reachable states of a bounded model, including a deliberate-breakage discipline that confirms the checker emits a shortest counterexample rather than silently passing. Its central invariant, that no action reaches an effector without a capability token issued by a control ring, is complete mediation and typed closure enforced together. Two of its reported results bear directly on claims made here. First, the invariant set survived a more than doubling of capability surface without being weakened or redesigned, which is empirical support for the general proposition that a declared contract boundary can absorb capability growth. Second, its measured governance overhead of 0.021~ms per request is small enough to remove latency as a prima facie objection to mandatory mediation; we treat that figure as a lower bound for our own boundary, because it covers invariant evaluation and not the schema validation, graph construction, and binding that a resolved Harness contract also requires, and we therefore make no overhead estimate of our own.

What that work does not settle is the question this paper asks. Its invariants govern the action-to-effector path inside one runtime; ours govern the capability-to-capacity binding between a Skill population and a Scaffold pool. Verification is over a bounded model rather than a deployed capacity range, and no capability-by-capacity factorial is run, so nothing in it estimates whether adding compatible Scaffold instances leaves semantic behavior invariant or whether adding capability leaves the capacity-response function unchanged. The gap between a verified safety core and a tested separability claim is precisely the space P1 occupies.

\paragraph{Skills as governed assets.}
Skillware supplies a software ontology for persistent, versioned behavioral artifacts and lifecycle continuity \cite{fan2026skillware}. SkillCorpus examines consolidation and evaluation at ecosystem scale \cite{wang2026skillcorpus}. Cross-layer misalignment research shows that a Skill description and its observed behavior can diverge \cite{zhang2026misalignment}, and a dedicated treatment of Skill security supplies the threat models, attack classes, and evaluation vocabulary that a semi-trusted Skill assumption requires \cite{badhe2026skillsecurity}. Together these sources support treating a Skill as a governed asset whose declared interface requires behavioral validation; they do not show that any particular Skill is safe, valuable, or portable. AEVAL separates first-attempt performance from self-correction and separates the executor from the grader \cite{anand2026aeval}. We carry those controls into the release and training protocols because business task success and runtime contract adherence must be reported as distinct evidence.

\paragraph{Harness evolution and structured change.}
Harness Handbook identifies behavior localization and editable structure as central problems in harness evolution \cite{wang2026handbook}. Continual optimizer evaluation finds that gains compound only when regression control remains inside the optimization loop \cite{wang2026compound}. GRACE structures persistent agentic context as a typed semantic graph and validates each proposed edit only within the local neighborhood of the nodes it touches, reporting a substantial reliability gain over an otherwise identical flat-text baseline \cite{hsu2026grace}. Two further systems arrive independently at a decomposed credit signal for harness or Skill repair: gated semantic quality-diversity separates model proposal from deterministic measurement and archives accepted patches by pathology \cite{luo2026gsme}, and workflow-localized mechanism learning attributes a failure to a workflow node and mechanism before choosing the smallest valid edit \cite{lin2026wml}. A further system makes the harness itself the object that accumulates experience, carrying lessons from past episodes forward rather than re-deriving them each run \cite{huang2026memoharness}; we take that as evidence that harness state is a governed asset, not as evidence that any particular accumulation policy is safe --- which is why Section~\ref{sec:memory-dualtrack} makes retention, deletion, and authorization contract decisions rather than architectural properties. A complementary negative result bounds how much any such loop can be trusted: harness optimizers have been observed proposing guardrails for failure classes that never occurred \cite{wang2026phantom}. These results motivate localized change, regression gates, and explicit intermediate structures. Sections~\ref{sec:intermediate-relation} and~\ref{sec:training-before-freezing} state what remains distinct in each case.

\paragraph{Code as the execution surface, and sandboxes as the capacity unit.}
A survey of code-driven agent systems argues that routing agent behavior through executable, verifiable, stateful code is the direction of travel for harness design \cite{ning2026codeharness}; that framing is the background against which our Skill-as-Code freezing step in Section~\ref{sec:skill-as-code} should be read, and it is why we treat frozen code as a determinism anchor rather than merely an efficiency device. On the capacity side, millisecond-level sandbox checkpoint and rollback makes the Scaffold instance a cheap, restorable unit rather than a heavyweight allocation \cite{dong2026deltabox}. That capability is directly relevant to the reset and washout requirements of Section~\ref{sec:evaluation}: a crossover design depends on returning a cluster to a comparable state between periods, and checkpoint/rollback is one mechanism by which the reset sentinels of that section could be met in practice. We cite it as an enabling mechanism, not as evidence that our sentinels are achievable.

\paragraph{Context scale and activation depth.}
Two contemporary studies constrain the evaluation of selective activation rather than the architecture itself. A controlled study of progressive disclosure reports that its benefit shrinks toward zero once a harness already partitions and retrieves competently, becoming decisive only at corpus scale, and that adding a second routing layer never helped \cite{he2026disclosure}. A white-box study of Skill execution under long contexts observes requirement coverage above 92\% while task outcomes collapse \cite{xue2026longcontext}. Neither result speaks to capability-to-capacity separability, but together they fix two controls that Section~\ref{sec:evaluation} must hold --- retrieval competence and activation depth --- and one metric it must report, the fraction of runs in which every declared requirement was satisfied rather than aggregate recall alone.

\paragraph{Source preservation and retrieval paths.}
Three recent studies constrain the data-substrate claim. In financial-document decision tasks, summary-only conditions increased disagreement relative to rereading the source, and restoring source context recovered only part of the lost performance; the result supports keeping canonical evidence reachable, not a universal theorem that every summary is harmful \cite{lee2026summaries}. At repository scale, lexical ranked retrieval overtook a raw file-navigation agent in the reported large-corpus regime and used far fewer query tokens, which motivates ranked discovery before expensive navigation but does not establish BM25 as universally optimal \cite{wang2026bm25scale}. On one 780-page report with 51 questions, an explicit read-and-navigate operator substantially outperformed the reported dense-retrieval baseline, while costing more and remaining statistically unresolved against BM25 \cite{tamang2026beyondtopk}. Together these results motivate a layered path --- ranked discovery, structured navigation or operators, then reading versioned source evidence --- and require the paper to distinguish online latency from amortized indexing and navigation cost.

\paragraph{Earlier design motivation.}
Earlier drafts used Policy-Driven Runtime Layer (arXiv:2605.27744) and Tool Forge (arXiv:2605.28000) as design leads. They are recorded here only as manuscript history; this revision asserts no result from either. The targeted August--September update does not replace the earlier and classical foundations needed to establish the novelty boundary.

\paragraph{Industry evidence, separated by plane.}
Contemporary industry sources illustrate the same evidence split without establishing the proposed architecture. An AWS account of agentic AI in enterprise resource planning supplies business/use-case evidence about workflow transformation \cite{aws2026erp}. AWS material on AgentOps and distributed hybrid-cloud agentic workloads supplies system/runtime evidence about operational governance and deployment topology \cite{aws2026agentops,aws2026hybrid}. None validates the complete Skill--Harness--Scaffold architecture, P1, the 0 to 100{,}000 design target, or the organizational thesis.

\paragraph{Recent executable systems and the remaining data boundary.}
Prime Agent integrates persistent computation, versioned long-term state, and recursive execution accounting, but its benchmark comparisons do not establish capability-capacity separability \cite{karten2026prime}. Apodex 1.1 combines environment and coordination scaling with evidence-based synthesis and controlled artifact delivery \cite{apodex2026}; those axes differ from capability versus compatible physical capacity. AutoResearch separates producing experimental artifacts from accepting claims about them \cite{ren2026autoresearch}. We adopt evidence-bound completion as a design requirement, not as a new invention or a reproduced reliability result.

VoiceMem separates upper-level memory organization from replaceable backend engines and reports index and routing ablations \cite{xie2026voicemem}; that evidence motivates separately measuring candidate discovery and contract enforcement, not importing speech-memory scores into enterprise data tasks. NeoHorse-1 explicitly separates predicted capability demand from the route actually served, which may reflect service availability or policy \cite{neohorse2026}. That distinction is a necessary control when measuring capacity interventions. None of these works establishes the specific combination of version-pinned 5W1H+Which predicates, execution-time revalidation, evidence binding, and the P1 protocol proposed here; conversely, this paper does not reproduce their trained models or benchmark gains.

\paragraph{Novelty boundary.}
Our contribution is therefore not a new reference monitor, capability model, modularity principle, scheduler, quality taxonomy, autonomic loop, retrieval algorithm, or the identification of the harness as the scaling locus. It composes established mechanisms into an agentic-runtime responsibility model, makes the data-use boundary executable through typed contextual predicates and version checks, and makes separability plus enforcement cost falsifiable. The combination and its testable consequences are the proposed contribution; the individual metadata dimensions, reference-monitor checks, and dependency traversal are established ideas. Table~\ref{tab:responsibility-boundary} states the ownership and non-ownership claims of the proposed architecture; it is not a completeness rating of products or organizations.

\begin{table}[t]
\centering
\caption{Responsibility boundaries in the proposed architecture. The table defines handoffs and non-ownership; it is not a completeness rating of products or organizations.}
\label{tab:responsibility-boundary}
\begin{tabular}{P{0.14}P{0.21}P{0.22}P{0.20}P{0.23}}
\hline
Boundary & Business semantics & Runtime governance & Physical execution & Enterprise data \\
\hline
Skill & owns & declares requirements & does not own & consumes governed contracts \\
Harness & compiles activated capability & owns & binds but does not supply & mediates access and evidence \\
Scaffold & does not interpret & exposes enforceable facts & owns & enforces locality and network boundary \\
Data substrate & preserves domain meaning & supplies policy/provenance facts & does not schedule & owns semantic integration and routing \\
\hline
\end{tabular}
\end{table}

\section{Responsibility Model and Primary Claim}
\label{sec:model}

The tuples and equations in this section specify the measurement contract rather than a full operational model. Section~\ref{sec:executable-ir} supplies a narrow operational account of data-use admission and execution; it does not formalize the entire runtime.

\subsection{Four Responsibility Objects}
\label{sec:runtime-objects}

\paragraph{Skill.}
A Skill $s\in\mathcal{S}$ is a versioned, independently deployable behavior declaration. It owns task intent, typed inputs and outputs, permitted effects, preconditions, tests, and release identity. Its minimum interface is $s=\langle g,i,o,p,e,v\rangle$, where $g$ describes the goal and applicability conditions, $i$ and $o$ are typed input and output schemas, $p$ contains preconditions and policy tags, $e$ declares externally visible effects, and $v$ identifies the validated version. A Skill may package instructions, tool adapters, executable code, examples, and verifier rules, but it does not choose physical placement, grant itself credentials, or redefine platform isolation. A Skill is \textit{operation-closed} when every external effect it can request is declared and testable through its interface. The unit used in this paper is an activated Skill bundle on an admitted path, not a passive registry entry.

Collection-scale evidence suggests that operation closure is not what current practice provides. The SkillMD-138K snapshot released with the Skillware ontology \cite{fan2026skillware} contains 138,133 content-deduplicated \texttt{SKILL.md} records drawn from 20,556 repositories. Frontmatter is nearly universal (136,380 records, 98.73\%), which the authors read narrowly as evidence that a structured metadata envelope is a dominant visible convention, explicitly not that its fields are validated or that activation semantics agree across agent systems. A lexical detector finds an explicit path token into \texttt{scripts/}, \texttt{references/}, or \texttt{assets/} in 32,069 records (approximately 23.2\%); the authors note that this detector misses implicit dependencies, so the figure is a lower bound rather than a census. Neither figure speaks directly to the fields operation closure requires: a frontmatter envelope establishes that a name and a description are conventional, corresponding to the goal field $g$, but not a typed output schema, a declared effect set, or a version identity that a gate could reject independently. That gap is the reason the Harness must enforce closure as an admission condition rather than assume it of registered Skills. Whether closure can be retrofitted onto existing Skills at acceptable cost is an empirical question we do not settle.

\paragraph{Harness.}
The Harness $\mathcal{H}$ is the runtime compiler and governor. Given a request $q$, identity and policy context $u$, and a Skill registry $\mathcal{S}$, it produces a candidate execution graph $D$, validates the graph, binds accepted nodes to Scaffold instances, and emits a trace:

\begin{equation}
(q,u,\mathcal{S}) \xrightarrow{\mathcal{H}} (D,C,b,\tau).
\end{equation}

Here $C$ is the resolved Harness contract, $b$ is a binding from graph nodes to Scaffold instances, and $\tau$ is the audit and replay trace. Harness owns logical admission and control: path construction, schema checking, policy evaluation, effect authorization, logical budgets, binding constraints, version identities, and evidence obligations. Selection and graph construction may use a language model and are therefore probabilistic. Admission, schema validation, policy evaluation, budget enforcement, effect authorization, and trace creation must be replayable for a fixed contract and policy snapshot.

\paragraph{Scaffold.}
A Scaffold instance $x\in\mathcal{X}$ is a physical execution envelope $x=\langle r,z,l,a\rangle$, where $r$ describes available resources, $z$ the isolation and trust zone, $l$ the data and network locality, and $a$ the attestation and runtime identity. A Scaffold may be a process sandbox, container, virtual machine, browser session, graphics processing unit (GPU) allocation, or remote execution pool. It supplies compute, process or container boundaries, network and data locality, runtime identity, resource accounting, queueing, failure containment, and attestation. It exposes resource facts to the Harness but does not reinterpret task goals or silently add capability semantics. As the execution and control boundary it owns these NFRs: performance, reliability, availability, observability, manageability, security, isolation, portability, elasticity, and capacity evolution.

\paragraph{Three Scaffold modules.}
The sandbox supplies the isolation boundary and the in-instance execution interface, including mounts, resource limits and exit reporting. API routing has two tracks. Model routing selects among eligible serving replicas. Tool routing enforces scoped destinations and records the outcomes of external operations. Platform governance manages leases, warm pools, quotas, artifact identity and telemetry. The Harness admits a Skill and authorizes each operation. The platform enforces those decisions at load and dispatch. Snapshot systems such as DeltaBox illustrate how physical restore costs can be reduced \cite{dong2026deltabox}. Restored state still requires a current tenant lease and valid execution identity. Table~\ref{tab:scaffold-controls} gives the observable interface for all three modules.

\paragraph{External data substrate.}
The external data substrate owns source authority, semantic definitions, snapshots, lineage, credentials, retention, and policy-relevant data facts. It is not a hidden Scaffold database and not a Skill-owned cache. The Harness resolves authorized access and evidence requirements; the Scaffold enforces the resulting locality and network constraints. Section~\ref{sec:data-substrate} specifies it.

These are responsibility boundaries, not mandatory teams or services. One implementation can combine objects, but its measurements must still distinguish logical admission, physical execution, and data authority. Information hiding is useful only when the hidden decisions have explicit interfaces and ownership \cite{parnas1972criteria}.

\subsection{Canonical Configurations and Outcomes}

The activated capability configuration $c$ identifies the exact versioned Skill bundle, admitted graph, tool and effect surface, model settings, and verifier set exercised by the workload. The Scaffold capacity configuration $s$ identifies the compatible worker count, resource class, isolation topology, placement constraints, and queueing limits. Capability is denoted by $c$; the symbol $s$ is reserved for Scaffold capacity in the experimental sections and is never used there for Skill.

The declared operating region $\Omega$ fixes the workload family and mix, tenant and identity model, Harness and policy versions, model and tokenizer versions, external-service contracts, data snapshots, compatible Scaffold classes, scheduler regime, and failure regimes. A result outside those bounds is a new study, not an extrapolation.

For each assigned cluster-period we observe three named quantities:

\begin{itemize}
\item runtime response $R(c,s)$: throughput, admission rate, queueing delay, latency distribution, saturation, retries, and cost per completed run;
\item semantic outcome $Q(c,s)$: task success, every-requirement satisfaction, output schema adherence, authorized-effect adherence, and postcondition satisfaction; and
\item enforcement overhead $E(c,s)$: added latency, compute, memory, control-plane work, evidence volume, and monetary cost attributable to admission, mediation, isolation, and measurement.
\end{itemize}

The literal names runtime response $R(c,s)$, semantic outcome $Q(c,s)$, and enforcement overhead $E(c,s)$ are part of the measurement contract. They prevent a favorable throughput result from substituting for semantic stability or a favorable semantic result from hiding an unaffordable control path.

\subsection{P1: Cost-Aware Separability}
\label{sec:p1}

Let $c_0$ and $c_1$ be preregistered capability configurations and let $s_0$ and $s_1$ be compatible Scaffold capacity configurations. The primary runtime endpoint $Y_R(c,s)$ is the natural logarithm of period-level p95 admission-to-terminal latency in seconds. Every admitted request is included; a request without a terminal event by the preregistered timeout $\tau$ is assigned latency $\tau$. Because requests assigned $\tau$ form a point mass at the censoring bound, and that mass can dominate the period-level p95 when timeout rates differ across arms, the report must also give the per-cell timeout rate and run a sensitivity analysis on an alternative endpoint (for example, the p95 of completed runs or a winsorized quantile); the equivalence decision on the primary endpoint must be robust to the recorded timeout distribution. The primary difference-in-differences estimand is

\begin{equation}
\Delta_R =
[Y_R(c_1,s_1)-Y_R(c_0,s_1)]
-[Y_R(c_1,s_0)-Y_R(c_0,s_0)].
\end{equation}

An illustrative two-sided equivalence margin is $m_R=\log(1.10)$, a 1.10-fold bound on the ratio of latency ratios: equivalently, an interaction of at most a 10\% multiplicative change in p95 tail latency. The runtime criterion is satisfied only when the 90\% two-sided cluster-aware confidence interval for $\Delta_R$ lies wholly inside $[-m_R,m_R]$, the two one sided tests (TOST) rule at level 0.05 \cite{schuirmann1987tost}. This is an equivalence test, not a directional non-inferiority test. For more than two levels, the same named endpoint, scale, margin, and interval rule apply to every preregistered capability-by-Scaffold interaction contrast in a simultaneous cluster-aware model. The other components of $R(c,s)$ remain reported secondary endpoints and cannot replace $Y_R$ after outcomes are observed.

P1 does not constrain the Scaffold main effect $Y_R(c,s_1)-Y_R(c,s_0)$. A positive Scaffold main effect on throughput, or a negative main effect on latency, is expected and does not falsify P1. The primary semantic endpoint $Q_{\mathrm{req}}(c,s)$ is the proportion of admitted runs satisfying every preregistered requirement. At each capability level, the capacity contrast is $D_Q(c)=Q_{\mathrm{req}}(c,s_1)-Q_{\mathrm{req}}(c,s_0)$. Semantic stability uses directional non-inferiority: with an illustrative $m_Q=0.05$, the one-sided 95\% cluster-aware lower confidence bound for every required $D_Q(c)$ must exceed $-0.05$ \cite{piaggio2012noninferiority}. Other semantic outcomes are secondary unless a separate scale, direction, margin, and multiplicity rule is preregistered.

The enforcement budget $B_E$ is fixed before outcomes are examined and applies to the vector of overhead measures: every mandatory component of $E(c,s)$ must remain within its declared latency, compute, evidence-volume, and monetary limits under the paired counterfactual in Section~\ref{sec:evaluation}. A cheap uninstrumented path cannot satisfy this budget because missing control work makes the evidence ineligible.

\begin{proposition}[Proposition P1: cost-aware capability-capacity separability]
\label{prop:p1}
Within the declared operating region $\Omega$, an admitted capability intervention preserves the log-p95 latency capacity-response relationship under the two-sided interaction-equivalence rule, a compatible Scaffold intervention preserves every-requirement satisfaction under the directional non-inferiority rule, and the measured controls remain within $B_E$.
\end{proposition}

\paragraph{Deriving every margin from a named decision.}
Because the design turns on margins, one rule binds all of them and not only the two stated above. Every equivalence, non-inferiority, and interaction margin must be derived from a stated decision that the margin governs --- the largest semantic degradation a release board would accept, the smallest capacity-response departure that would change a procurement or placement choice --- and never from the observed variance of a pilot, which would let a noisy instrument certify itself by widening its own margin. Record the decision, its owner, and the resulting numeric margin as separate preregistered fields so that a reader can dispute the decision without recomputing the statistics. This rule applies equally to the reset sentinel bounds of Section~\ref{sec:evaluation}, which must be supplied together with decision-derived values for the primary endpoints before any study is registered. The 10\% and five-percentage-point examples above are not approved operational tolerances.

\subsection{The Four-State Decision Rule}
\label{sec:decision-rule}

The verdict has four states. The fourth exists because the six obligations of Section~\ref{sec:obligations} form a conjunction, and in a real enterprise runtime the most likely outcome of a first study is that some obligations clear their thresholds and others do not. Treating the conjunction as the only admissible entry point would make a study that discovers one undeclared edge report nothing at all about P1, which is not an acceptable design.

\begin{itemize}
\item \textbf{Supported within $\Omega$.} All six obligations pass coverage, calibrated-sensitivity, and worst-case violation-bound gates; the 90\% interaction interval lies inside $[-m_R,m_R]$; every semantic lower bound exceeds $-m_Q$; and every enforcement-overhead upper bound lies below its component of $B_E$.
\item \textbf{Falsified within $\Omega$.} Eligibility and calibration are sufficient for a decision, and the interaction interval lies wholly above $m_R$ or below $-m_R$, a semantic upper bound lies below $-m_Q$, an enforcement-overhead lower bound exceeds its budget.
\item \textbf{Conditional-engineering result.} Coverage and calibration are adequate, but one or more obligation violation lower confidence bounds exceed their preregistered ceilings. The study reports the outcome estimands in full, together with the named obligations whose rates exceeded threshold and the observed value of each, and carries no verdict on P1 in either direction.
\item \textbf{Inconclusive.} Instrumentation coverage or monitor calibration is insufficient, an interaction, semantic, violation, or overhead interval does not resolve its margin, crossover contamination remains after reset, a predeclared eligibility rule fails, or another required quantity cannot be estimated with the planned cluster-aware uncertainty.
\end{itemize}

A conditional-engineering result characterizes the engineering state of the runtime under test --- which contract obligations it can and cannot currently hold --- rather than the scientific status of P1. It is publishable and informative: it identifies which obligations are expensive to establish in a real runtime, and the observed rates supply the dose axis for a later study of how outcome departure scales with residual coupling. What it must not do is enter the P1 decision, because an outcome measured under a known open channel cannot discriminate between separability and the channel. A study that ends in this state should report the engineering cost of closing each violated obligation, since that cost is the practical barrier to testing P1 at all.

The rule is applied in this order: insufficient coverage, calibration, reset, or estimability yields inconclusive; otherwise a clearly violated obligation yields conditional-engineering; unresolved obligation bounds yield inconclusive; only after all obligation gates pass are the P1 outcome and cost contrasts classified as supported, falsified, or inconclusive. This precedence prevents an open-channel case from simultaneously being called falsified and conditional-engineering. The rule also distinguishes failure of the claim from failure to measure it. In particular, a nonsignificant interaction from an underpowered design is inconclusive, not support.

\subsection{Proposition Inventory and Evidence Planes}

P1 is the core structural claim. The proposition identifiers are those of this line of work rather than a consecutive series: P2--P14 were stated in earlier drafts and have since been withdrawn, absorbed into the six obligations, or demoted to mechanism hypotheses. The three that remain separately testable keep their original numbers so that the protocols and the earlier record stay comparable: P15 (dual-subgoal gate benefit, Section~\ref{sec:training-before-freezing}), P16 (task-family evidence-path sufficiency) and P17 (bounded policy-contract decoupling, Section~\ref{sec:intermediate-relation}).

\begin{longtable}{P{0.22}P{0.15}P{0.32}P{0.31}}
\caption{Claim types and evidence status. ``Proposed experiment'' denotes a protocol, not a completed study in this paper. The first two rows are definitional and interpretive choices rather than testable claims; for them the relevant question is whether the distinctions are useful and consistently applied, not whether an experiment confirms them.}
\label{tab:propositions}\\
\hline
Object & Claim type & Scoped claim & Evidence status \\
\hline
\endfirsthead
\multicolumn{4}{c}{\tablename\ \thetable{} (continued)}\\
\hline
Object & Claim type & Scoped claim & Evidence status \\
\hline
\endhead
\hline
\endfoot
\endlastfoot
Enterprise-runtime responsibility architecture & Design proposal & Four responsibility objects separate business capability, runtime governance, execution/NFR ownership, and enterprise data governance through stable contracts and change cadence & Architectural synthesis and finite data-use and Scaffold fixtures (\S\ref{sec:ir-conformance}, \S\ref{sec:scaffold-binding}); deployment unevaluated \\
Shared organizational contract & Interpretive thesis & Architecture makes communication, ownership, change cadence, and evidence obligations explicit across enterprise groups & Definitional framing choice of this paper, not a testable prediction; adoption value unevaluated \\
P1 & Proposition & Cost-aware capability-capacity separability inside $\Omega$ & Analytic argument; proposed experiment (\S\ref{sec:evaluation}) \\
Path safety / mediation & Proposition & Complete mediation enables path-level admission checks & Analytic argument; cited external evidence; proposed experiment (\S\ref{sec:secondary}) \\
P15 & Hypothesis & Output-plus-process gates improve discrimination, attribution, or training & Cited external evidence; proposed experiment (\S\ref{sec:secondary}) \\
P16 & Hypothesis & A pinned address/index policy with source fallback supports task-family reconstruction & Proposed experiment (\S\ref{sec:secondary}) \\
P17 & Proposition & Compatible source and output changes remain bounded by versioned policy contracts & Analytic argument; cited external evidence; proposed experiment (\S\ref{sec:secondary}) \\
Write-time representation foundation & Analytic result & Query-agnostic deterministic representations are sufficient exactly when their induced partition preserves every distinction required by the declared task family & Self-contained finite-alphabet derivation (\S\ref{sec:formal-data-foundations}); no empirical validation \\
Dry-run / locality & Hypothesis & Planning cost is offset by avoided work or movement & Proposed experiment (\S\ref{sec:secondary}) \\
Skill lifecycle / training & Design pattern & Train, validate, freeze, monitor, and roll back Skills & Cited external evidence; proposed experiment; future work \\
Prefix stability / seed affinity & Design pattern & Tool-set cohesion and same-seed co-location improve cache reuse and locality & No protocol in this paper; measurement is future work \\
Amdahl and handoff calibration & Open question & Serial fraction and orchestration efficiency are linear in tool-set overlap and handoff cost & No protocol in this paper; requires a design that manipulates overlap and handoff size \\
Higher-order memory & Open question & Structures beyond the contract-relevant IR & Future work \\
\hline
\end{longtable}

\begin{table}[t]
\centering
\caption{Two required evidence planes for enterprise adoption. Each plane answers a different decision question; neither plane substitutes for the other.}
\label{tab:evidence-planes}
\begin{tabular}{P{0.17}P{0.25}P{0.32}P{0.26}}
\hline
Evidence plane & Primary decision question & Representative evidence & Does not establish \\
\hline
Business/use-case & Does the versioned Skill produce the intended business outcome for its declared users and workflow? & Task and requirement completion, output quality, cycle time, human correction, adoption, and outcome-specific benefit or loss & Runtime admission integrity, isolation, capacity, availability, portability, or cost behavior \\
System/\allowbreak runtime & Does the admitted path execute within its declared contract and enterprise NFR envelope? & Gate and trace integrity, path and effect conformance, throughput, active-agent capacity, latency, saturation, failures, isolation, availability, backpressure, and cost & Usefulness, workflow fit, user acceptance, or business value \\
\hline
\end{tabular}
\end{table}

The business owner and Skill team require the first plane; runtime governance, platform, SRE, security, operations, and data authorities require the second. A release decision must name the evidence required from both planes and must not use success in one as a proxy for the other.

\section{Contract Boundary and Architecture}
\label{sec:architecture}

\subsection{The Resolved Harness Contract}
\label{sec:harness-contract-def}

For an admitted run, the Harness emits a resolved contract

\begin{equation}
C=\langle I,O,G,A,B,V\rangle,
\end{equation}

where $I$ and $O$ are the run-level typed inputs and outputs, $G$ is the activated Skill graph, $A$ is the authority and effect set together with its evidence obligations, $B$ contains resource, time, token, cost, concurrency, placement, and isolation constraints, and $V$ pins policy, model, registry, data-snapshot, capability, verifier, and trace identities needed for replay. The boundary's interface semantics are defined by five observable components: the admission decision over $C$, the activated path in $G$, the authorized effects in $A$, the binding constraints in $B$, and the postconditions encoded by $O$ and its evidence obligations. Each field is serializable, inspectable, and independently rejectable. A run rejected at admission produces a reason and no external effect. An accepted run carries the contract identity through execution and evidence.

\begin{figure}[t]
\centering
\includegraphics[width=\linewidth]{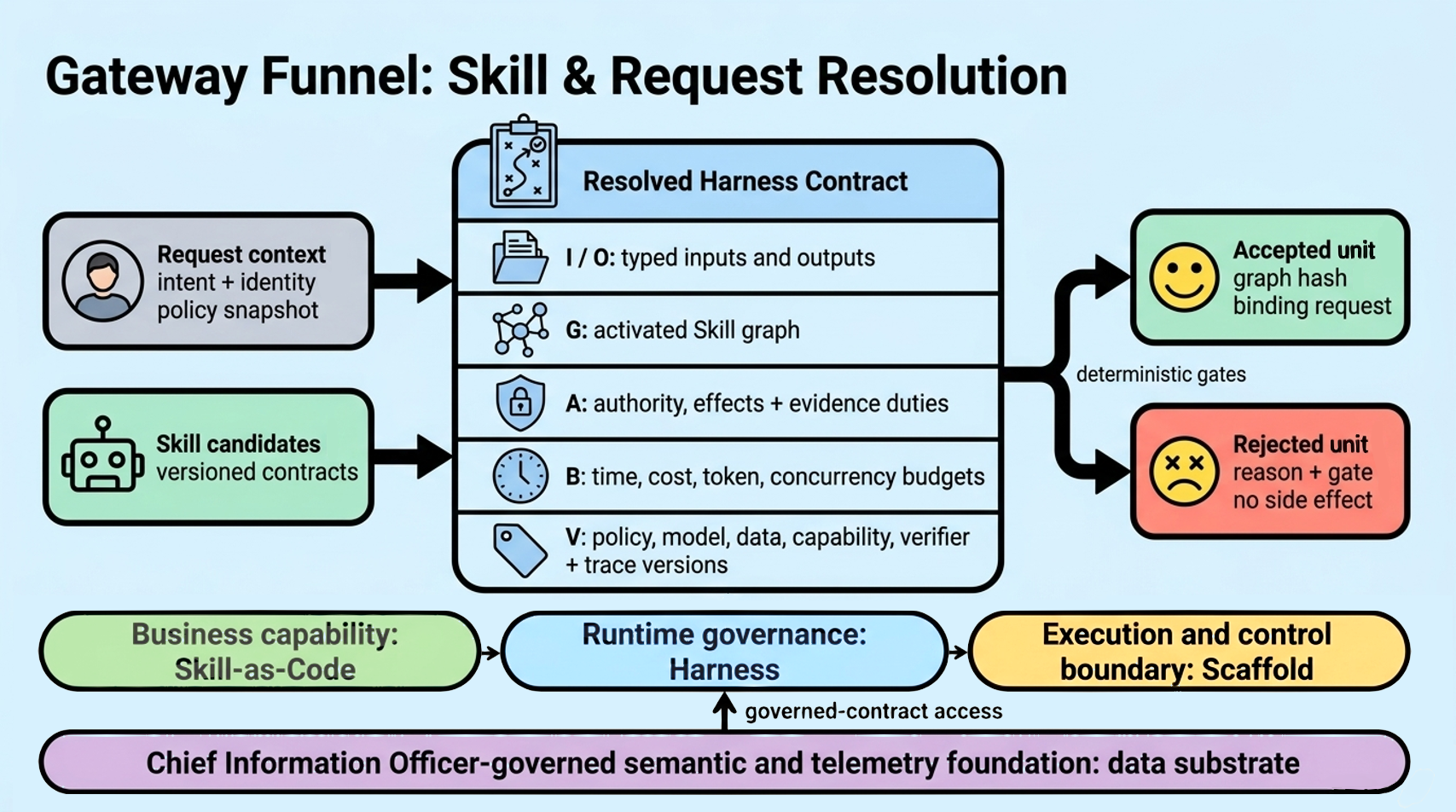}

\caption{Shown: a request and activated bundle pass deterministic admission gates to become a resolved Harness contract with typed data, path, authority, budgets, versions, binding, and trace identity. Why it matters: the contract is the inspectable unit on which complete mediation, replay, and capability-to-capacity binding are tested. Class: architecture.}
\label{fig:harness-contract}
\end{figure}

Figure~\ref{fig:harness-contract} makes the Harness boundary concrete. A useful shorthand is: \textit{probabilistically written, deterministically and auditably executed}. The phrase does not imply that execution outcomes are deterministic; external services and models remain variable. It means that the authorization decision, declared effects, selected versions, and evidence requirements can be reconstructed from the recorded contract.

The contract also makes selective activation observable. The registry can contain many Skills, but $c$ changes only when activated behavior on an admitted path changes. This prevents registry cardinality from being mistaken for a capability intervention and gives the experiment a stable treatment identity.

\subsection{When may a Scaffold replace a worker?}
\label{sec:scaffold-binding}

Suppose a monthly reconciliation run reads two ledger snapshots and publishes one discrepancy report. Adding a worker in the same trust zone can reduce queueing time. A worker that reads a newer ledger, serves a different tokenizer, or issues a second publication changes the task. A zero process exit code cannot distinguish these cases. The Harness therefore specifies which physical substitutions preserve the admitted node, and the Scaffold returns evidence about the worker that actually ran it.

\paragraph{The dispatch decision.}
For each node $n$ of $G$, project $A,B,V$ onto the authorized tool operation and identity scope, the hard placement constraints and resource floor, and the pinned execution profile. At dispatch time $t$, compare these with the observed $r,z,l,a$ of candidate $x$. Let $H_C(n,x,t)$ mean that the isolation zone, locality, current lease epoch, unexpired attestation, loaded artifact digest, permitted model profile and authorized tool route all match. The model profile pins model weights or a service revision, tokenizer, prompt template, decoding policy and serving precision. An authorized tool route identifies the destination and scoped principal. Replicas may have distinct physical identities. Each must supply evidence for the same declared profile. Given a current admission and an unexpired run budget, the decision is
\[
 d_C(n,x,t)=
 \begin{cases}
 \mathrm{reject} & \text{if a hard constraint fails or its evidence is missing},\\
 \mathrm{defer} & \text{if }H_C(n,x,t)\text{ holds but capacity is insufficient},\\
 \mathrm{bind} & \text{if }H_C(n,x,t)\text{ holds and capacity is available}.
 \end{cases}
\]
Rejection concerns this candidate. The Scaffold may try another candidate within the admitted class. Changing the profile requires fresh Harness admission. Deferral consumes the same deadline. Queue release, snapshot restore, and failover each require a fresh decision. In a distributed implementation, the lease and reservation must cover the transition to actual execution, or the effector must fence stale dispatches. An endpoint that cannot substantiate the contracted model identity is ineligible for this comparison. Matching a profile establishes eligibility only. Output quality still has to meet the semantic non-inferiority margin.

\begin{table}[htbp]
\centering\small
\caption{Six Scaffold functions and the observation that makes each inspectable. The Harness sets the constraint and evaluates business postconditions.}
\label{tab:scaffold-controls}
\begin{tabular}{P{0.23}P{0.37}P{0.40}}
\hline
Function & Physical enforcement & Evidence for the run \\ \hline
Isolation & Allocate the declared tenant and trust zone. & Instance, zone, mount scope and lease epoch. \\
In-sandbox execution & Load the pinned executable and bound its resource use. & Measured image and code digests, exit status and output handle. \\
Model routing & Select a replica serving the admitted model profile. & Requested and served profiles, replica identity and failover event. \\
Tool routing & Enforce destination, credential scope and effect recovery policy. & Route, scoped principal, operation key and authoritative receipt. \\
Capacity and lifecycle & Reserve slots, enforce tenant quota and revalidate restored workers. & Wait and start times, warm or cold state, lease and release. \\
Artifact and telemetry governance & Verify publisher and artifact at load, then join physical events to the node. & Verification identity, observed digest, timestamps and correlation identifiers. \\
\hline
\end{tabular}
\end{table}

Table~\ref{tab:scaffold-controls} can guide implementation without requiring a single product. OpenSandbox exposes lifecycle, execution and egress interfaces \cite{opensandboxRuntime}. Kubernetes Agent Sandbox exposes templates, claims and warm pools \cite{k8sAgentSandbox}. Envoy documents per-route retry and timeout controls \cite{envoyRouter}. Using these components leaves the comparison between the admitted profile and the observed executor to the integrating runtime.

\paragraph{Write timeout and execution receipt.}
A timed-out publication may already have committed. Before any second attempt, resolve the same tenant, logical operation, payload digest and request identifier against an authorized status source. A committed receipt completes the physical operation. A terminal no-commit result must exclude any later commit by the first attempt. If status is unknown, a retry is admissible only if the effector atomically deduplicates the same key and payload across all candidate replicas for the whole retry horizon. Otherwise hold the operation for reconciliation. Every new dispatch also requires a current binding and remaining budget. Receiving a past receipt requires read authority and creates no new publication. The final trace joins the contract and node, selected and loaded code, worker and lease, route and effect receipt, and resulting artifact. For the ledger example, the business verifier must still reconstruct totals and check the disposition of unmatched entries. SkillLifeBench documents how the selected skill and loaded code can diverge \cite{zhang2026skilllife}. That distinction motivates collecting the execution-side digest independently.

\paragraph{Executable controls.}
The ancillary Scaffold fixture compares a single admitted profile with concrete worker observations. Eight binary mutations change slots, zone, region, lease epoch, expiry, loaded digest, tokenizer or tool destination. The test oracle classifies cases from the injected labels. The implementation inspects the resulting values. Across all 256 combinations, one bound, one deferred for lack of capacity and 254 rejected. A separate control changed worker identity and increased capacity while retaining eligibility. Ten named write-recovery cases checked the decision after a timeout, including a committed write, an unknown outcome without shared deduplication, and an expired lease before retry. The fixture trusts observations and status responses. It exercises neither a live executor nor a distributed effector.

\paragraph{Measurement in the crossover.}
Randomize worker count only within the declared compatible classes. Retain every assigned request in its cluster-period, including deferrals, rejections, and unresolved effects at the observation deadline. Record time to an accepted result, verified completion per assigned request, and the marginal cost of revalidation and reconciliation. Reporting only successful rebindings would hide the cost of a stricter gate. Independently marked diagnostic cases can force stale lease epochs, model-profile mismatches, or ambiguous write acknowledgements to calibrate the monitors. Section~\ref{sec:evaluation} defines the separate outcome analysis.

\subsection{Control Plane and Data Plane}
\label{sec:manageability}

The control plane owns Skill registration, contract compilation, policy versions, placement rules, and release state. The data plane carries request payloads, executes accepted nodes, moves authorized evidence, and emits traces. The planes interact through references and resolved contracts rather than by copying the full control state into every model prompt.

\begin{figure}[t]
\centering
\includegraphics[width=\linewidth]{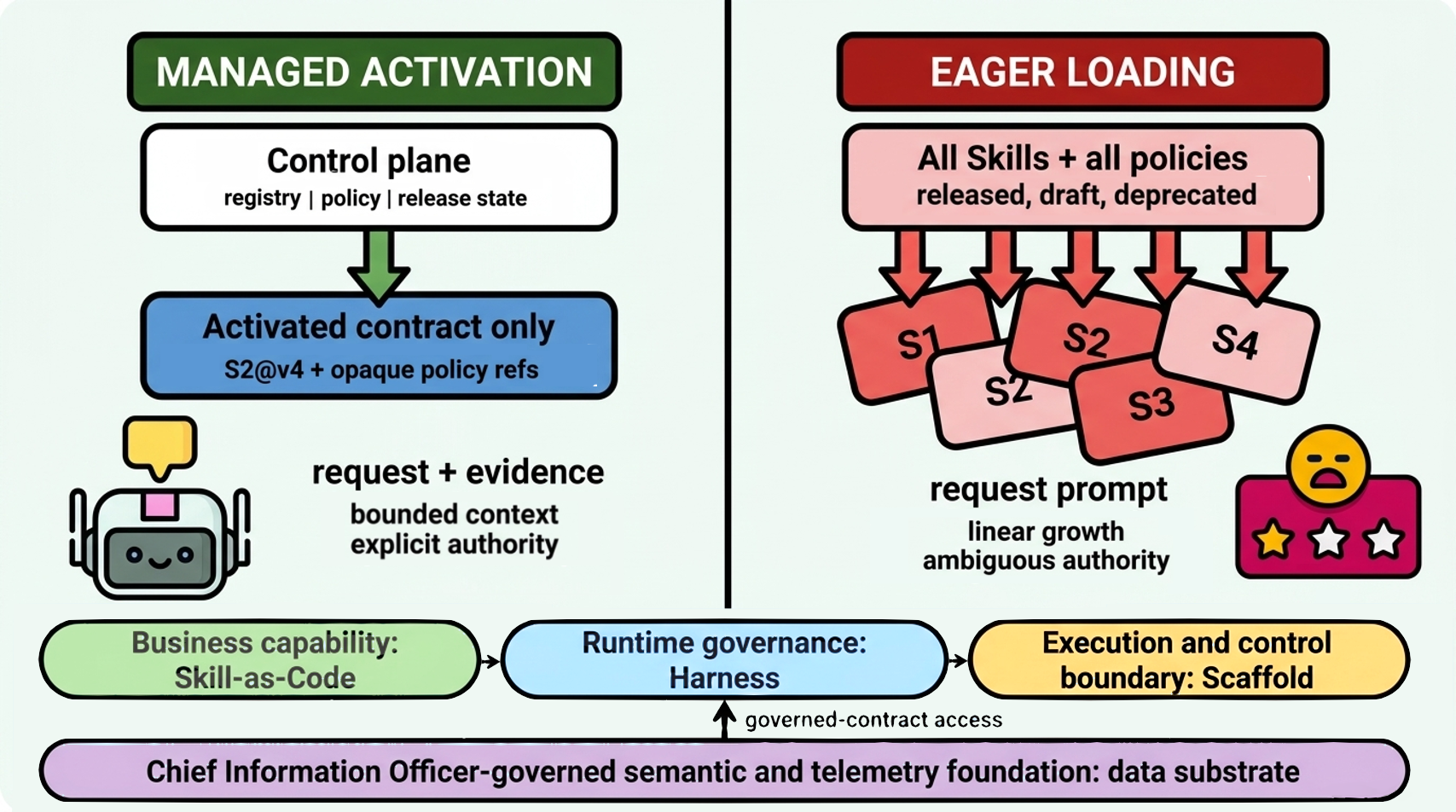}

\caption{Shown: the control plane holds the registry, policy, admission, and placement logic, while the data plane receives only the resolved contract and executes accepted work; the leaking variant injects every schema and policy into the request path. Why it matters: keeping inactive registry state and mutable control metadata off the request path limits unmeasured coupling between capability growth and runtime load. Class: architecture.}
\label{fig:control-data}
\end{figure}

Figure~\ref{fig:control-data} highlights a failure mode that is easy to mistake for flexibility. Loading every Skill, application programming interface (API) schema, and policy example into each request leaks control-plane state into the data plane. Besides token cost, the leak creates ambiguous authority: a description visible to the planner may look executable even when its release or policy state has changed. The Harness should first select a small candidate set from registry metadata, then materialize only the accepted versions and constraints needed by the run.

Separation is not established by drawing two planes. The complete-mediation and scheduler-independence obligations must observe every crossing. A cache fill, quota check, retry controller, or placement hint that bypasses the resolved contract remains a possible coupling channel.

\subsection{Path-Aware Invariants}

Per-Skill approval is necessary but insufficient. Let $G_q=(V_q,E_q)$ be the activated graph for request $q$, with nodes representing Skill operations and edges representing data or control dependencies. The Harness evaluates invariants over the path:

\begin{equation}
\mathrm{Accept}(G_q)=\bigwedge_k \phi_k(G_q,u,A,B,V).
\end{equation}

Examples include separation of duties, data-residency continuity, prevention of read-then-publish paths, maximum cumulative privilege, and mandatory human approval before an irreversible effect. A path can violate an invariant even when each node is locally allowed. Runtime checks must also cover dynamically inserted nodes and retries, since they change the effective graph.

\subsection{Audit, Replay, and Reversibility}

An audit trace should distinguish three questions: what the planner proposed, what the deterministic gates accepted, and what the bound Scaffold actually executed. Conflating these stages makes incident review nearly impossible. At minimum, the trace records request and principal identity, candidate and activated Skills, contract versions, policy decision inputs, graph hash, binding decisions, external calls, evidence artifacts, and final effects.

Replay has levels. \textit{Decision replay} recomputes gate outcomes from recorded inputs. \textit{Plan replay} reruns planning under the recorded model and prompt version, accepting that stochastic output may differ. \textit{Execution replay} re-executes against pinned data or service snapshots when available. The runtime should declare which level it supports rather than promising exact replay across mutable external systems.

Contemporary work audits agent execution online to predict failure early from observable signals \cite{zhang2026agentforesight}. That capability is complementary to, and not a substitute for, what Section~\ref{sec:obligations} requires: predicting that a run is likely to fail is a different problem from establishing that a monitor observed every event in a declared channel set. An early-warning signal can be accurate while coverage of the threat model remains unmeasured, which is why the obligations below demand an independent event inventory rather than a predictor's confidence.

Manageability also requires a stop path. Skill releases can be disabled, policy versions rolled back, bindings revoked, and in-flight graphs cancelled at Harness checkpoints. These mechanisms turn a trace from passive observability into operational control.

\subsection{Threat Model}
\label{sec:threat-model}

The Harness is a reference monitor, so what it protects and against whom must be stated before the six obligations can be read. We assume the following, and each assumption is a place the architecture can be attacked rather than a property it establishes.

\begin{itemize}
\item \textbf{Trusted computing base.} The control plane is trusted: the Harness compiler, the Skill registry, the policy evaluator, the gate implementations, and the Scaffold attestation service. A compromise of any of them defeats the boundary, which is why the design requires replicated control services, signed artifacts, and independent attestation, and degrades by reducing capability rather than widening governance.
\item \textbf{Untrusted model output.} Planner and executor output is an untrusted request for authority, never an authorization. A plan that names an operation outside the resolved contract is a rejected request, not a widened contract; this is why the admission decision is deterministic for a fixed contract and policy snapshot while plan generation is not.
\item \textbf{Semi-trusted Skills.} A Skill author is authenticated and accountable but not assumed correct or benign. A Skill may declare one behavior and exhibit another, and contemporary work reports exactly that divergence \cite{zhang2026misalignment} and catalogues the resulting attack surface \cite{badhe2026skillsecurity}. Declarations are therefore admission inputs to be validated behaviorally, and operation closure is a gate rather than a claim.
\item \textbf{Untrusted external content.} Tool results, retrieved documents, and data-source payloads are attacker-influenceable. Text that arrives through the data plane cannot change the activated path, the authorized effect set, or a policy version, because those are resolved in the control plane before the fetch occurs; contract resolution, rather than instruction filtering, is the response to prompt injection.
\item \textbf{Adversary capabilities.} The adversary can author or modify a Skill, influence request text, control external content, and compose approved Skills into a hazardous path. The adversary cannot break Scaffold isolation, forge attestation, or subvert the control plane; those are Scaffold and platform security properties assumed rather than argued here.
\end{itemize}

The design goal is narrow: every effect an adversary can reach must appear in some resolved contract's authorized effect set, and every attempt outside that set must be a typed failure recorded in the trace. Gates and traces make a violation attributable, and complete mediation makes an undeclared channel detectable, but neither prevents an authorized effect from being harmful. Deciding which effects should be authorized remains a business and policy judgment outside this architecture.

\section{Cohesion, Context Partitioning, and Derivation Closure}
\label{sec:cohesion}

\subsection{First Principle: The Agent's Job Is Organizing Context}
\label{sec:first-principle}

For a fixed model, fixed sampling or decoding configuration, fixed runtime tool configuration, and fixed external state, a single inference varies with its context input. Within that controlled setting, context is the variable an agent loop directly organizes for the next inference step. Across arbitrary agent runs the model, decoder, runtime tools, and external systems may also vary, so context is not literally the only variable. Planning, data retrieval, tool invocation, and memory writes nevertheless remain decisions about what to place in the next context and what to keep external to it.

This observation has a direct consequence for performance under serving systems that expose prefix-cache reuse. Cache identity requires more than semantically equivalent prompts: the model and tokenizer versions, system and policy metadata, tool serialization and ordering, and every prefix token must remain fixed. If a single token changes in the middle of the prefix, all subsequent positions must be recomputed. The reusable part is not ``the context that changed slightly'' but ``the context whose prefix is frozen token-by-token with changes only at the tail.'' Heterogeneous tools or goals can therefore reduce cache reuse and introduce irrelevant context, but this is one candidate mechanism for long-horizon degradation rather than a complete explanation.

\subsection{Context Partitioning by Tool-Set and Data Source}

The corresponding design heuristic is: \textit{place version-pinned stable material (system prompt, policy metadata, serialized tool schemas, and fixed memory) at the front in deterministic order; place per-turn task content at the tail.} Under a prefix-caching serving stack, a longer token-identical prefix creates a larger reuse opportunity. Stable semantic content alone is insufficient if serialization, ordering, tokenizer, or model versions differ.

One useful axis for partitioning context is the tool-set. A sub-agent that repeatedly uses the same version-pinned set of tools and Harness contracts can maintain a stable prefix when its prompt renderer is deterministic. Conversely, switching among heterogeneous tool-sets may rewrite that prefix and dilute attention. Stable tool membership therefore helps prefix stability but does not guarantee it; the system must measure token identity and cache hits directly.

The secondary axis is data. Detail such as raw tables, full document text, and verbose intermediate results should not flow between sub-agents as context. Instead, canonical detail is written to the data substrate, and a bounded evidence bundle is passed forward: declared conclusions, source and snapshot handles, evidence spans, key constraints, and unresolved obligations. Downstream sub-agents retrieve detail on demand through the governed discovery--navigation--read path rather than treating a natural-language summary as authoritative. Under explicit interface assumptions that evidence bundles are size-bounded, detail remains external, and downstream retrieval traffic and cost are accounted for separately, the handoff payload changes from $O(\text{full context})$ to $O(\text{evidence bundle})$. This is a conditional communication-size claim, not an unconditional reduction in total system work or information loss.

\paragraph{Production evidence for a tool taxonomy.}
The tool-set partitioning heuristic acquires empirical shape from a production-scale characterization of coding-agent traces --- 761M model calls and 775M tool invocations across 45 distinct tools in one week \cite{agenticcoding2026wild}. Usage is heavily concentrated: the top 11 tools account for over 90\% of invocations, and \texttt{get\_file} alone for 35.0\%. The observable tools fall into four performance classes whose system consequences differ. \textit{Read/retrieval} tools complete in tens of milliseconds with near-universal success and constitute the performance base of the loop. \textit{Mutation} tools cluster at 0.25--0.6~s, bounded by filesystem operations. \textit{Execution} tools (\texttt{run\_command\_in\_terminal} at 17.0\% of invocations, \texttt{run\_build}, \texttt{run\_tests}) are the latency and failure source: mean durations of 68--78~s against medians of seconds, success rates near 73\%, failed invocations taking $48\times$ longer at P95 than successful ones, and failed builds injecting 7--8$\times$ more tokens of diagnostics; 9\% of turns enter failure-driven retry that consumes roughly $4\times$ the compute of the median turn. A fourth, small \textit{meta/orchestration} class is high-frequency and near-zero-failure, and a 4.6\% long tail consists of customized tools, which is the empirical reason admission control must offer a registration channel rather than a fixed inventory. Two design consequences follow for the Sandbox module of Section~\ref{sec:runtime-objects}: read-class tools must remain sub-100~ms, while execution-class tools must execute asynchronously with timeout fuses and reclaimable resources. The trace is coding-domain; class shares and tail factors in other domains are an open measurement question.

\subsection{Four Mechanisms for High Cohesion and Loose Coupling}

The partitioning principles above yield four design mechanisms distributed across the three runtime responsibility objects:

\begin{enumerate}
\item \textbf{Tool-set cohesion (Skill).} A sub-agent should use one version-pinned tool-set and deterministic prompt serialization throughout a run. This increases the opportunity for prefix stability and cache reuse, which must still be measured. The anti-pattern is an agent that changes tool schemas, ordering, or policy metadata across successive turns without recording the resulting cache discontinuity.
\item \textbf{Minimally overlapping partitioning (parallelism).} When a task is decomposed into parallel sub-agents, their tool-sets should be as disjoint as task semantics allow. Disjoint tool declarations are only a proxy for a smaller shared surface: compute pools, locks, external services, authorization backends, data dependencies, and scheduler state may still create contention or correlated decisions and must be measured separately.
\item \textbf{Evidence-bundle handoff (loose coupling).} Sub-agents pass bounded conclusions together with source, snapshot, span, verifier, and pending-obligation handles, not full context. The receiving sub-agent retrieves authoritative detail on demand from the data substrate. Because that substrate is stack-external rather than a further runtime object, this absorption does not introduce its storage schema into the runtime core.
\item \textbf{Seed-affinity placement (Scaffold).} Each type of sub-agent has a derivation seed, an image with pre-warmed tools, dependencies, and cache. Same-seed sub-agents are co-located to reuse image layers, tool processes, and cache hot regions; different-seed sub-agents are spread across resource domains to reduce contention.
\end{enumerate}

The parallelism argument of mechanism 2 can be stated as a calibration hypothesis. If a fraction $\varsigma$ of a workload remains serial, Amdahl's law caps the speedup at $1/(\varsigma+(1-\varsigma)/N)$ for $N$ parallel sub-agents. We hypothesize $\varsigma=\varsigma_0+\lambda\rho+\eta z$, where $\rho$ measures tool-set overlap and $z$ collects independently measured shared-compute, lock, service, authorization, data, and scheduler coupling. We likewise hypothesize $E_{\mathrm{orch}}=E_0-\mu\bar{c}$ for orchestration efficiency and average handoff cost $\bar{c}$. Neither linear relation is implied by the architecture, and neither is tested by any protocol here: the crossover design of Section~\ref{sec:evaluation} varies capability and capacity but does not sweep $\rho$ or $\bar{c}$ as designed factors. We therefore register both calibrations as open questions, and note what a study would need: a design in which tool-set overlap and handoff evidence-bundle size are the manipulated variables at fixed workload and Scaffold pool, with the linear form itself under test against a monotone-but-nonlinear alternative.

Mechanisms 1 and 4 target intra-agent locality; mechanisms 2 and 3 target inter-agent coupling. Their evidence status differs and should not be blurred. Mechanisms 1 and 4 have no protocol here; their claims concern token-prefix identity, cache-hit rate, and co-location interference, which the present protocols do not manipulate.

\subsection{Derivation-Closure Orchestration}
\label{sec:derivation-closure}

The four mechanisms describe how a partition should look once chosen. They do not say how the runtime arrives at a partition for a request whose required tools and sources are not known in advance. We close that gap with a rule we call \textit{derivation closure}: each admitted sub-agent contract is closed over named operations, evidence sources, input/output and handoff schemas, effects, and budgets. Any reference outside that admitted closure is not satisfied in place but becomes the admission condition for deriving a new sub-agent.

\paragraph{Bounding the output range.}
Fixing those admitted operations, evidence sources, schemas, effects, and budgets bounds the sub-agent's authorized action-and-evidence range because each contract dimension is enumerable at admission time. It does not bound every token sequence the model might emit, nor does it guarantee semantic correctness. This is the operational reading of high cohesion: the sub-agent is cohesive not merely because its prompt is focused but because its executable authority and evidence duties are contractually closed.

\paragraph{References as derivation conditions.}
When a sub-agent encounters a needed search tool or data source that its contract does not cover, granting it inline would erode both the frozen prefix and the closure property. Instead the reference is recorded as an unmet dependency and triggers derivation: the Harness resolves a fresh contract whose tool-set and data-set cover that dependency, seeds a new sub-agent from the matching template, and binds it as a child. Capability growth within a single request therefore proceeds by adding bounded sub-agents rather than by enlarging any existing one. Each derivation event is an ordinary contract compilation and passes the same deterministic gates, so the request-level graph remains typed and replayable no matter how the partition was discovered.

\paragraph{The main agent as a satisfaction controller.}
The main agent does not execute task tools. Its function is to assemble returned evidence bundles and propose whether the run may terminate. For each output sub-domain $d$, the contract binds a versioned verifier $V_d$, its evidence inputs, and a threshold $\tau_d$ before execution. The controller performs a governed join across returned evidence bundles, applies top-$k$ selection to retain the best-supported evidence per field, and reports the \textit{satisfaction ratio} of the declared output sub-domains:

\begin{equation}
\mathrm{sat}(\sigma^{out})=\frac{1}{|\mathrm{dom}(\sigma^{out})|}\sum_{d\in\mathrm{dom}(\sigma^{out})} \mathbb{1}\big[V_d(\mathrm{join}_k(\text{evidence bundles})_d)\geq\tau_d\big].
\end{equation}

When every $V_d$ is deterministic or version-pinned with a preregistered error tolerance, the ratio can participate in a contract-level termination predicate. When a verifier is model-based, mutable, or uncalibrated, the ratio is an estimate and cannot by itself authorize termination; the contract must require an independent deterministic gate or named human approval. Subject to that distinction, the loop continues while the ratio is below its target and some unsatisfied sub-domain still admits an untried derivation. The remaining cases --- exhausted derivation budget or no candidate --- are typed outcomes whose unsatisfied sub-domains and verifier versions appear in the trace.

\begin{figure}[t]
\centering
\includegraphics[width=\linewidth]{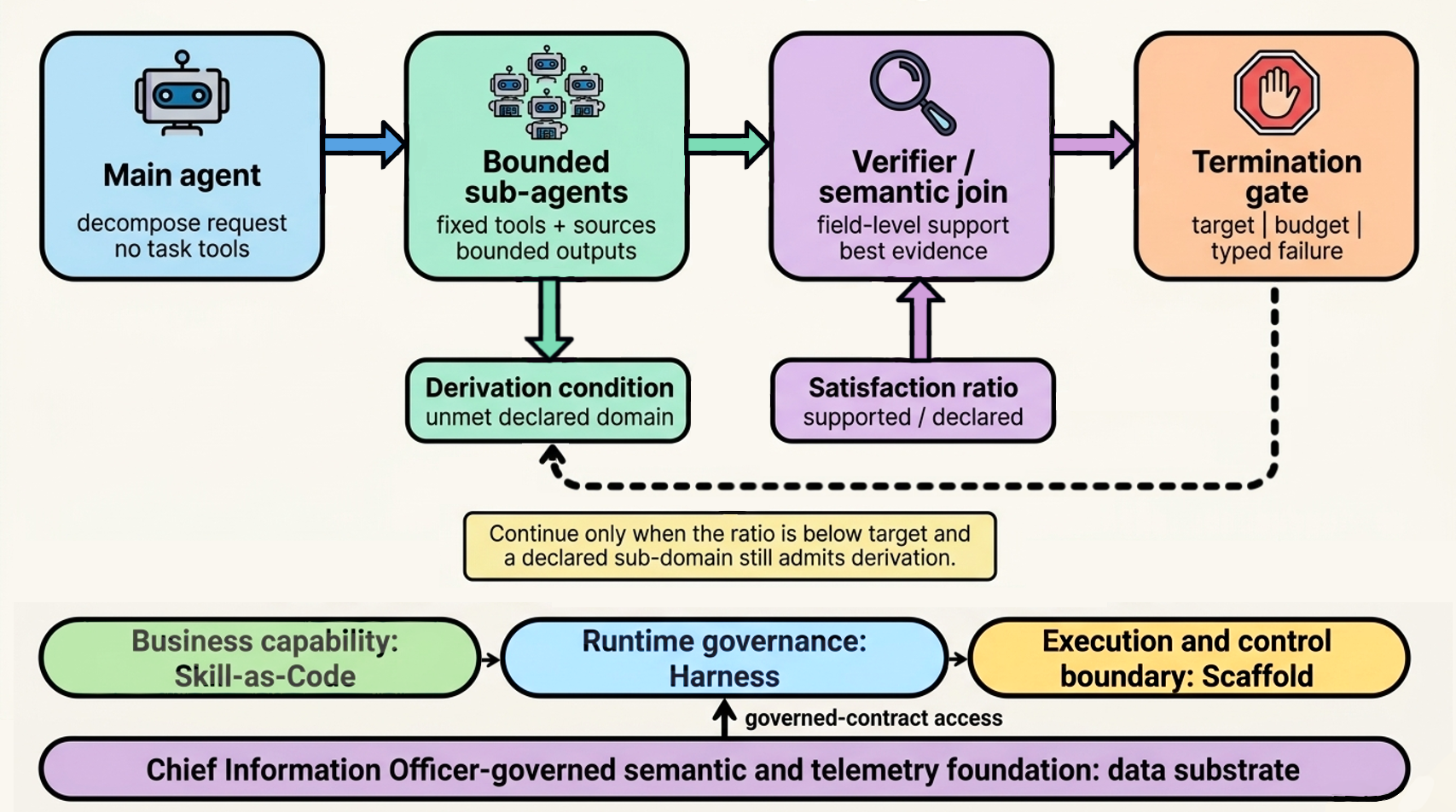}

\caption{Shown: a main agent derives bounded sub-agents for unmet dependencies, then a separate verifier or semantic join checks their returned evidence before a termination gate closes the path. Why it matters: capability expansion during a run remains a sequence of admitted contracts rather than an invisible increase in ambient authority. Class: protocol.}
\label{fig:derivation-closure}
\end{figure}

Figure~\ref{fig:derivation-closure} shows the loop. Three properties are testable. First, version-bound verifier results can move termination from an implicit model judgment into an inspectable contract predicate, subject to verifier validity. Second, because derivation is the only way to widen tool or data access, the trace is a complete record of admitted and activated capabilities, not a claim about every capability the task counterfactually needed. Third, the per-sub-domain structure aligns orchestration with the training gate of Section~\ref{sec:training-before-freezing}, allowing the same declared decomposition to be evaluated at training and run time.

\subsection{Parallelism, Locality, and Planning-Time Dry-Run}

Once the Harness has a graph, it can expose parallelism that a monolithic prompt obscures. Independent nodes may execute concurrently; dependent nodes remain ordered by graph edges and effect constraints. The Scaffold pool contributes locality and capacity facts. The binding algorithm can then co-locate data-intensive nodes, isolate high-risk effects, and reserve scarce accelerators only for nodes that declare them.

A planning-time dry-run resolves graph structure, candidate bindings, budget envelopes, policy decisions, and expected evidence without committing external effects. It answers operational questions before execution: which data regions will be crossed, which nodes can run in parallel, which irreversible effects require approval, whether predicted cost is within budget, and whether all required Scaffold types are available.

\begin{figure}[t]
\centering
\includegraphics[width=\linewidth]{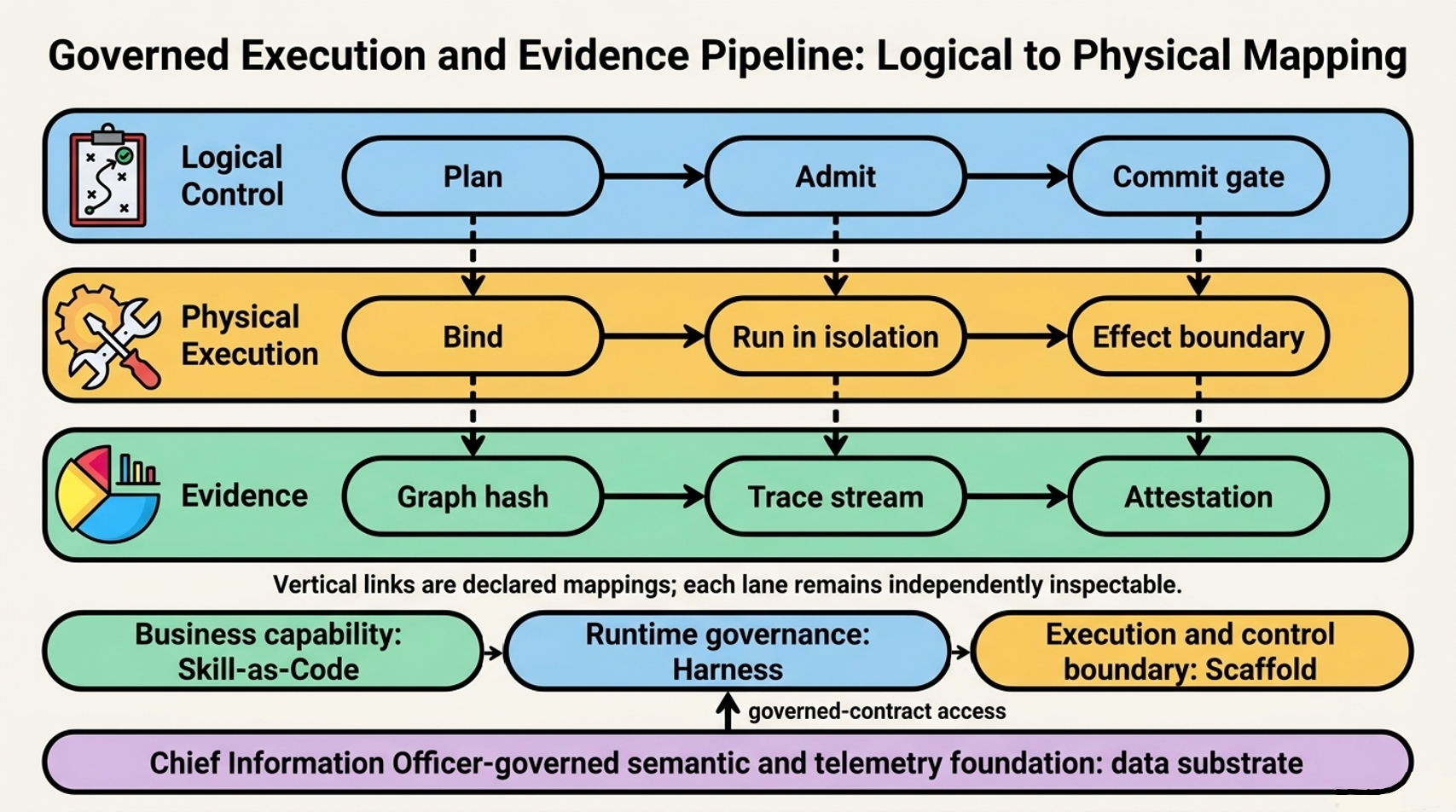}

\caption{Shown: logical control validates and binds a graph, physical execution runs it in compatible isolation and locality zones, and an evidence path records both decisions and effects; the vertical links are illustrative, potentially many-to-many contract mappings rather than a one-to-one logical-to-physical correspondence. Why it matters: placement optimization remains distinguishable from semantic policy while still exposing the resource channels that could recouple capability and capacity. Class: architecture.}
\label{fig:dry-run}
\end{figure}

Figure~\ref{fig:dry-run} supports a modest claim: graph visibility creates an optimization opportunity; it does not guarantee a speedup. Dry-run is not assumed to be beneficial. Its latency and avoided-work value are part of enforcement overhead $E(c,s)$, and locality can conflict with load balance, policy, or accelerator availability. A study can use immediate admission instead, but it must still record equivalent binding and rejection facts.

\section{The External Data Substrate}
\label{sec:data-substrate}

\subsection{External Data as a Contract Instantiation}

External data should not be a privileged side channel into the agent. It is a Harness-contract instantiation with explicit authentication, authorization, schema, snapshot, provenance, freshness, residency, and evidence obligations. A data Skill may discover or query assets, but the resolved contract determines what can be fetched and what evidence must accompany the result.

\begin{figure}[t]
\centering
\includegraphics[width=\linewidth]{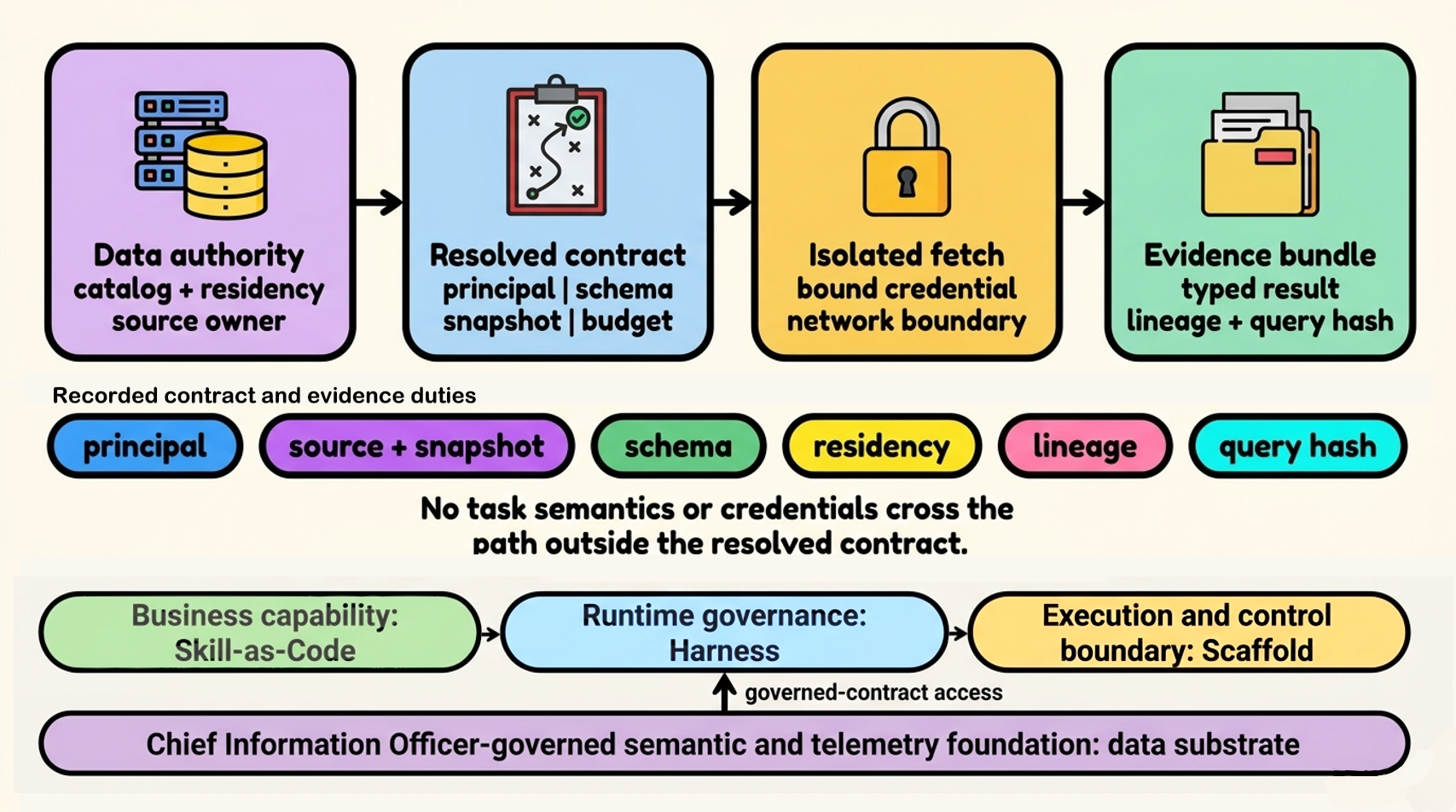}

\caption{Shown: the data authority resolves a semantic request into a governed contract, the Scaffold performs an isolated fetch, and the run receives an evidence bundle rather than source credentials; the figure's shorthand means that no \emph{undeclared} task semantics or credentials cross the boundary, while declared request fields and evidence obligations necessarily do. Why it matters: data policy and snapshot identity remain fixed experimental inputs instead of hidden properties of the capability or capacity treatment. Class: architecture.}
\label{fig:external-data}
\end{figure}

Figure~\ref{fig:external-data} shows how this boundary preserves both portability and accountability. The same analytical Skill can run against different warehouses or document stores when their adapters satisfy the contract. Conversely, the same data source can support multiple Skills without learning their task semantics. Evidence returned to the model is a typed bundle carrying source and snapshot identity, not an unqualified text fragment. Within $\Omega$, the data snapshot and policy-relevant facts are fixed or explicitly assigned as a failure regime; a live source whose contents drift without version identity makes semantic non-inferiority inconclusive.

\paragraph{Hypothetical AI4Science walkthrough.}
Consider a request to screen perovskite candidates for a target band gap near 1.3~eV using recent literature and to produce a cited, reproducible report. The choice of a research workload rather than a procurement, customer-service, or finance case is deliberate. Those three are the cases where the contract boundary is easiest to draw, because their data sources, effects, and approval authorities are already institutionally settled. Corporate research is the harder case on every dimension the architecture claims to handle: the source set is heterogeneous and partly external, the compute requirement is specialized enough to test whether capacity really is substitutable, the effects include an irreversible outward publication, and the evidence obligation is a per-claim citation rather than a workflow outcome. Resolving this walkthrough would not establish transfer to those other cases; each workload requires its own contract and evaluation. The walkthrough is an architecture walkthrough, not an implemented case study.

The explicit chain is \textit{intent $\rightarrow$ contract compilation $\rightarrow$ path gate $\rightarrow$ binding $\rightarrow$ trace $\rightarrow$ typed failure $\rightarrow$ recoupling}. At \textit{intent}, the request fixes the band-gap target together with recency, citation, and reproducibility constraints. During \textit{contract compilation}, the Harness activates literature-review, materials-screening, and report-generation Skills and closes their typed inputs, outputs, effects, and evidence obligations. The \textit{path gate} checks data authorization, institutional high-performance-computing identity, a network allowlist for literature and materials databases, and effect closure before any fetch or computation. \textit{Binding} resolves approved databases, a microVM for scripts, and a compatible accelerator without exposing infrastructure credentials to the Skills. The \textit{trace} pins Skill and model versions, literature and materials-data snapshots, binding and scheduler decisions, screening-script identity, and citation evidence. A missing authorization, ungrounded citation, unit anomaly, or absence of compatible capacity is returned as a \textit{typed failure} rather than repaired through an undeclared path. Finally, \textit{recoupling} is visible if a shared quota, scheduler feedback, or resource-sensitive scientific output makes a Scaffold change alter the admission decision, activated path, authorized effects, binding constraints, postconditions, or scientific result.

\paragraph{Field-level sketch of the resolved contract.}
Because the value of a contract boundary depends on whether its fields can actually be written down, we make the same walkthrough concrete at field granularity. The intent above resolves to one $C=\langle I,O,G,A,B,V\rangle$ whose fields carry, illustratively: $I$ names \texttt{target-property} = band gap, \texttt{target-value} = 1.3~eV, \texttt{tolerance} = 0.1~eV, \texttt{chemistry-family} = perovskite, \texttt{literature-window} = 24 months, and \texttt{citation} = required. $O$ is a typed report with three output sub-domains: \texttt{candidate-table} (composition, predicted gap, uncertainty, method), \texttt{literature-synthesis} (one resolvable citation per asserted claim), and \texttt{reproduction-bundle} (script hash, environment digest, input snapshot identifiers). $G$ is the activated three-node graph \texttt{literature-review} $\rightarrow$ \texttt{materials-screen} $\rightarrow$ \texttt{report-compose}, with the second node's inputs typed to the first node's citation-bearing output. $A$ authorizes \texttt{read:literature-db}, \texttt{read:materials-db}, \texttt{execute:screening-script}, and \texttt{write:report-artifact}, while \texttt{write:external-publication} is absent and therefore refused; its evidence obligations are \texttt{per-claim-citation}, \texttt{unit-check-passed}, and \texttt{script-hash-recorded}. $B$ carries \texttt{gpu-class} = A100-or-compatible, \texttt{max-gpu-hours} = 40, \texttt{max-tokens} = 2M, a \texttt{cost-ceiling}, \texttt{concurrency} = 8, \texttt{isolation} = microVM, and \texttt{data-residency} = institutional. $V$ pins the Skill versions, planner and executor model identifiers, policy snapshot, literature and materials-database snapshot identifiers, binding record, and trace root hash.

Three of these fields carry the walkthrough's load. The absence of \texttt{write:external-publication} from $A$ is what turns an attempted outward post into a typed failure instead of a silent success. The \texttt{tolerance} field in $I$ is what makes \texttt{candidate-table} scoreable by a deterministic verifier, which is the precondition for the satisfaction ratio of Section~\ref{sec:derivation-closure} to carry any weight. And the pair of snapshot identifiers in $V$ is what separates a reproducible disagreement between two runs from an unexplained one. The field names are ours and illustrative rather than normative.

\subsection{From Query Plan to Intermediate Relation}
\label{sec:intermediate-relation}

The stack-external data substrate preserves canonical raw or source-native records together with versioned snapshots, provenance, access policy, and deterministic read or extraction operations. An asynchronous background loop may build indexes, summaries, proposed relations, and navigation structures, but these are auditable derivatives rather than the sole authority. Here we make the ``plan'' step concrete as an explicit, versioned data-use policy we call the \textit{Intermediate Relation} (IR), situated between a source-oriented address/index catalog and an output-oriented contract registry. The abbreviation is unfortunately overloaded in adjacent literature, where IR more often denotes an \textit{intermediate representation} --- a compiler-style artifact that a generator emits and a backend consumes. Our IR is not that: it is a versioned, auditable \textit{relation} between a task theme and the source set admitted for it, and it is read by the Harness at admission. The executable ticket introduced below is a separate, request-bound artifact derived from this relation; it does not change the meaning of IR.

\paragraph{The Intermediate Relation.}
The \textit{Data Wiki} is an address/index/operator catalog over those authoritative records. Each entry identifies the source, snapshot or validity interval, provenance, format, access boundary, deterministic read or extraction operations, retrieval terms, typed cross-source relations, and any derived natural-language description together with its producer, version, confidence, and evidence spans. The operational path is layered: global ranked discovery narrows the candidate set; hierarchical navigation or compiled operators resolve fields and relations; and the agent reads the versioned source evidence needed for the decision. A summary may improve discovery or planning, but it cannot be the only retained representation and its presence does not imply task-family sufficiency.

Section~\ref{sec:formal-data-foundations} states the formal reason for this boundary without assuming that every summary fails. For a fixed finite task family, a deterministic write-time representation is exactly sufficient only when its induced partition preserves every source distinction needed by every admitted query. Enlarging the future query family can only refine that minimal sufficient partition and weakly increase its entropy. A coordinate-query construction then separates a $k$-bit query-agnostic sufficient representation from a $ceil(log_2 k)+1$ bit read-time request-and-return path when the versioned source remains addressable. The comparison deliberately separates persistent representation cost from online communication cost; storage, indexing, authorization, freshness, and latency remain explicit system costs. P16 therefore tests bounded task-family evidence-path sufficiency with governed fallback, not universal sufficiency of an index or summary.

\begin{proposition}[Hypothesis P16: task-family evidence-path sufficiency]
\label{prop:p16}
For a preregistered target report family and field set, a version-pinned Data Wiki address/index policy with governed source fallback supports held-out reconstruction equivalently or non-inferiorly to a token-matched direct-reading policy, while exceeding metadata-only, shuffled-index, and informative field-ablation controls. The claim is relative to that task family, access budget, source snapshot, and verifier. It does not assert that an index or summary preserves every possible meaning, and raw retention alone does not guarantee accessibility, freshness, correctness, or authorization.
\end{proposition}

The \textit{Theme Wiki} holds a versioned task-family output and quality contract: required fields, report structure, verifier versions, evidence granularity, uncertainty treatment, freshness rules, and allowed fallback or escalation. Templates for reports, charts, decks, analytical write-ups, or data-model programs are registered only after the corresponding output sub-domain has been validated under the lifecycle of Section~\ref{sec:training-before-freezing}. The two registries are intended to change independently only through bounded interfaces: a source catalog should not encode every presentation demand, and an output contract should not bind to one physical source.

\paragraph{Bounded policy coupling.}
The IR is the declared coupling point between them, not a proof that no hidden coupling exists. For a given theme it records source identifiers and snapshots, allowed discovery and navigation operations, evidence requirements, fallback rules, compatibility ranges, and seven contextual dimensions:
\begin{equation}
\mathrm{IR}:\ \mathrm{theme} \mapsto \langle \{(\mathrm{src}_i,\mathrm{snapshot}_i,\mathrm{ops}_i)\},\ \mathrm{5W1H}\mathrm{+}\mathrm{Which},\ \mathrm{evidence},\ \mathrm{fallback},\ \mathrm{compatibility} \rangle,
\end{equation}
where \textbf{When} states validity and update cadence; \textbf{Where} states the access-domain boundary; \textbf{Who} states ownership and authorization; \textbf{What} states business semantics; \textbf{Why} records a declared integration rationale or causal hypothesis rather than hidden chain-of-thought; \textbf{How} states the allowed physical or logical access operations; and \textbf{Which} states typed temporal, lineage, and cross-reference relations. Each derived field carries confidence, provenance, evidence spans, and a schema version. These dimensions are metadata and policy inputs, not substitutes for source evidence. With the IR explicit, intent selects a Theme contract; ranked discovery finds candidate sources; the IR constrains navigation, evidence, and fallback; the Harness validates freshness and authorization; and the Scaffold executes only the admitted reads or operators.

To make the IR concrete at field granularity, an illustrative entry for the theme \texttt{quarterly-revenue-report} is: sources = \{\texttt{sales-system}@\texttt{2026Q1-close}, \texttt{billing-system}@\texttt{2026Q1-close}\}; allowed operations = \{\texttt{rank}, \texttt{read-table}, \texttt{read-json}, \texttt{join-on-product-line}\}; When = validity window 2026Q1, refreshed daily at 02:00; Where = intranet; Who = finance-analyst group only; What = revenue by product line per accounting close; Why = a declared reconciliation hypothesis requiring both accounts-receivable and point-of-sale sources; How = the named lakehouse table and sales-system JSON adapter; Which = typed cross-references to the product-line dimension, the 2026Q1 accounting-close event, and the lineage edge to \texttt{finance.revenue\_raw}; evidence = row identifiers, source spans, snapshot identifiers, and verifier version; fallback = escalate unresolved mismatches to the data steward. The request first discovers the two sources, then navigates only through admitted operations, and finally reads the cited rows or JSON objects. A stale or unauthorized hit is rejected before any fetch, and neither source's full schema is copied into the agent context.

\begin{proposition}[Proposition P17: bounded policy-contract decoupling]
\label{prop:p17}
Within preregistered compatibility ranges and in the absence of hidden shared state, a change compliant with one registry's versioned interface does not require an unanticipated schema or module change in the other registry. Incompatible semantic changes, evidence-policy changes, or newly exposed dependencies must update the IR and may require coordinated migration; those expected changes do not count as decoupling failures.
\end{proposition}

The proposition is conditional on bounded interface complexity, substitutable implementations, explicit version compatibility, admission control, and no undeclared shared state. It concerns unanticipated propagation, not immutability. A source or output change may legitimately modify entries, tests, generated artifacts, and the IR; the falsification question is whether a change declared compatible causes propagation outside the preregistered dependency graph.

\paragraph{Contribution boundary.}
The general principle that explicit structure can confine a change's blast radius has independent support in the scoped-verification work discussed in Section~\ref{sec:related} \cite{hsu2026grace}. The IR applies that principle to the coupling between a source-oriented catalog and a task-family output contract. This subsection proposes no new retrieval or data-integration algorithm. Its contribution is the systems-level organization of authoritative sources, derivative indexes, operators, and data-use policies into an auditable boundary; retrieval quality, source accessibility, policy completeness, and propagation bounds remain unproven.

\subsection{Executable 5W1H+Which Contracts}
\label{sec:executable-ir}

The remaining gap is operational: a catalog can correctly describe a source while a cached decision permits an expired or revoked read. We therefore separate the declarative Intermediate Relation from its request-bound execution ticket. The relation remains a data-use policy; the ticket binds that policy to one principal, source snapshot, output contract, and policy epoch. Neither natural-language relevance nor a populated metadata field grants authority.

\paragraph{A motivating counterexample.}
A quarterly revenue task discovers two tables with identical column names. One reports consolidated USD revenue for the current quarter; the other reports local-currency subsidiary revenue for the previous quarter. Both contain a product identifier. A syntactically valid join can produce a plausible but invalid report. The What and When predicates must match business units and fiscal periods, Which must identify the approved join relation and cardinality, Who and Where must authorize the read and output destination, and How must restrict the executable operation. Why names an approved integration rationale; it is not a model-generated proof that the join is valid. Even if these predicates hold during planning, authorization may be revoked while a queued read waits for capacity. Reusing the old decision silently turns capacity-induced delay into a semantic and policy change.

\paragraph{Three independently versioned objects.}
A Data Wiki entry stores a source identifier, immutable snapshot identifier, content digest, typed field and relation descriptors, validity interval, policy reference, and available operations. A Theme Wiki entry stores the task-family identifier, required output fields, units and periods, verifier identity, allowed evidence granularity, and fallback policy. An IR entry connects those objects with the seven contextual dimensions and explicit dependency identifiers. Its human-readable explanations are reviewable metadata. Executable predicates and their trusted inputs determine acceptance.

For request $q$ from principal $u$, let $d$ be a source descriptor, $t$ the Theme contract, and $p$ the trusted policy snapshot. Admission resolves an IR entry into
\begin{equation}
J=\langle q,u,d,t,p,\mathrm{ops},\mathrm{budget},\mathrm{epoch},\mathrm{evidence}\rangle.
\end{equation}
The request identifier and principal are platform-supplied. Source, policy, and Theme versions are immutable references, not mutable names such as ``latest''. The ticket carries permitted operations and evidence obligations but no source credential. A production implementation must bind it cryptographically or hold it inside the trusted computing base; an ordinary serialized dictionary is not a capability token.

\paragraph{Check, bind, read, validate.}
The proposed boundary performs five steps. First, ranked discovery returns candidate source identifiers; relevance changes candidate order but does not bypass policy. Second, resolution obtains the authoritative descriptors and evaluates the When, Where, Who, What, Why, How, and Which predicates against the Theme contract. Third, the boundary binds the accepted versions and policy epoch into $J$. Fourth, immediately before each read or effect, the adapter rechecks expiry, revocation, budget, operation, and snapshot identity at a declared linearization point. Fifth, the returned evidence envelope binds bytes or spans to the source digest, operator version, units, period, and contract identity. Output validators check required fields against that envelope. A content hash establishes byte identity; it does not establish truth or semantic entailment.

An unsuccessful step returns a typed reason such as SOURCE\_CHANGED, STALE, UNAUTHORIZED, SEMANTIC\_MISMATCH, RELATION\_MISMATCH, OPERATION\_DENIED, or REVALIDATE. Fallback may select another already permitted source and obtain a fresh ticket. It cannot silently relax the fiscal period, unit, access domain, or evidence requirement. If a task genuinely requires different semantics, it creates a new contract revision. A request that cannot satisfy its output contract returns an explicit incomplete result or abstention according to the registered Theme policy.

\paragraph{Revocation semantics.}
The check and the read must share a linearization point with updates to the relevant policy and source state. In the in-process reference model this is a lock. In a distributed deployment it requires a declared transactional, leased, or epoch-fenced adapter protocol; a check followed by an unrelated network call is insufficient. A read linearized before revocation may already have released bytes. The architecture cannot retract those bytes, and downstream publication requires its own current authorization. Our guarantee concerns reads linearized after revocation, not retroactive deletion from an agent's context.

\begin{proposition}[Conditional soundness of the data-use boundary]
Assume trusted, authentic source and policy descriptors; complete mediation of every source read; a check-and-read operation atomic with respect to relevant updates; and evidence envelopes constructed from the bytes actually read. Then every successful mediated read satisfies the bound operation, principal, validity, semantic descriptor, relation, and budget predicates at its linearization point, and its envelope identifies those bytes and the checked versions. Proof: the boundary returns success only after all predicates hold under the same state used for the read. Atomicity excludes an intervening state change; envelope construction uses the returned bytes. The statement follows for each successful operation and therefore for a finite sequence by induction. It does not prove that the descriptors are correct, that all external channels are mediated, or that a model's final answer follows from the evidence.
\end{proposition}

\paragraph{Change invalidation and its limits.}
Let the dependency graph contain source snapshots, schema versions, policies, operators, IR entries, Theme contracts, and registered output artifacts. An edge points from an input object to a dependent object. For a changed set $U$, invalidate the reachable dependent set $I(U)$ and revalidate before reuse. A graph traversal costs $O(|V_U|+|E_U|)$ in the visited subgraph, excluding the cost of revalidation. This is standard dependency maintenance rather than a new graph algorithm. The proposed contribution is the declared set of policy, evidence, and semantic dependencies that the runtime must track.

If dependency capture is complete and each validator reads only its declared immutable dependencies, an object outside $I(U)$ has unchanged validator inputs. Its deterministic validation decision is therefore unchanged. This conditional locality statement does not establish P17 in a deployed system. Hidden SQL views, shared prompts, model updates, or mutable source aliases can violate dependency completeness. A conservative registry-wide epoch safely invalidates more objects; a fine-grained graph may reduce revalidation work but needs an independent dependency inventory. The reference model below deliberately uses the conservative epoch, so it does not claim the graph optimization has been implemented.

\paragraph{Evidence-bound completion.}
A file existing at the expected path is not sufficient evidence of completion. The output contract binds the artifact digest, required fields, evidence handles, verifier version, and unresolved obligations. The final transition distinguishes completed-and-verified, partial, rejected, and expired artifacts. A cached artifact whose source or policy dependency has changed cannot be delivered as newly verified. This requirement is especially relevant when a long-running branch finishes after a user intervention or policy update. It adapts established delivery and provenance mechanisms to the Theme/IR boundary rather than claiming that verified artifact delivery is itself new.

\subsection{Reference Model and Finite Conformance Study}
\label{sec:ir-conformance}

The supplementary artifact contains a standard-library Python reference model of one trusted source registry, an immutable contract, admission tickets, a registry-wide epoch, and an atomic check-and-read operation. It implements read authorization and evidence construction only. It has no language model, retrieval engine, multi-source join, distributed scheduler, sandbox, or output-publication service. The single registry lock supplies the model's atomicity assumption; it is not a measured scalable concurrency design.

We enumerated the Cartesian product of ten binary fault dimensions over one valid revenue-record fixture: source digest, Theme version, principal, domain, fiscal period, unit, approved rationale identifier, relation identifier, operation, and byte budget. Each dimension has a valid value and one predeclared invalid mutation. The independent finite-domain oracle accepts exactly the unmutated fixture. All $2^{10}=1{,}024$ cases are included, including multiple simultaneous faults; the gate reports the first rejection reason. The result is one accepted case, 1{,}023 rejected cases, and zero disagreements with the declared oracle. A CSV records every case and a JSON summary records the source-code digests and interpreter version.

Five lifecycle tests also pass: revocation after preparation invalidates the ticket; changed bytes under the same source version fail digest validation; expiry between preparation and execution rejects the read; repeated reads of the unchanged snapshot return the same envelope; and compatible revalidation retains the bytes while updating the epoch. These tests expose failures that a presence-only metadata validator would miss. They are specification conformance checks, not evidence of retrieval quality, adversarial robustness, or production reliability.

This is a finite exhaustive result for a constructed domain. The large fraction of invalid cases is an enumeration property, not an estimated prevalence of bad enterprise requests. Zero disagreement is not a confidence bound on future failures. No comparator framework is rated from these cases, no throughput or latency improvement is reported, and P1, P15, P16, and P17 remain unevaluated empirically. The contribution is an inspectable implementation of one boundary and a reproducible counterexample domain, narrowing the gap between prose contracts and executable obligations.

\subsection{A Study That Could Establish the Data Contribution}
\label{sec:data-study}

The confirmatory data study must separate relevance from admissibility. Use independently annotated report families with held-out organizations or source collections, held-out time periods, and natural as well as injected schema and policy changes. Split by source and template lineage before building the Data Wiki; random question splits can leak the same report structure into both development and evaluation. The annotator records required sources, units, temporal scope, authorized principals, join keys and cardinality, and sufficient evidence spans. Multi-source tasks must include both answerable joins and cases in which the correct action is refusal or clarification.

Compare tuned dense retrieval, BM25, hybrid retrieval, agentic source navigation, a source-preserving catalog with prose-only metadata, and the executable contract boundary. For the central causal contrast, give both catalog arms identical candidates, source-access tools, budgets, backbone, and descriptions; vary only whether the contextual fields are enforced at the access boundary. A second factorial contrast varies ranked discovery versus navigation while holding enforcement fixed. This prevents a gain from additional retrieval calls from being attributed to governance. Ablations remove individual predicates, source fallback, execution-time revalidation, or evidence binding. These ablations run only against safe fixture sources.

Primary outcomes are every-required-field correctness with valid provenance, unauthorized or stale release rate against an independent event inventory, and false refusal on valid tasks. Report answer quality, coverage, calibrated abstention, and conversion-limited cases separately: refusing every request is policy-conservative but useless. For numerical tasks, the oracle normalizes declared units and checks operands; it must not silently repair incompatible fiscal periods or entity scope. If a learned judge is needed for semantic entailment, calibrate it on blinded human labels and publish disagreements. A valid citation address alone is not entailment.

Report online latency and tokens together with catalog construction, incremental maintenance, source rereads, policy checks, revalidation work, and human annotation cost. Show total lifecycle cost at multiple reuse counts rather than calling background work free. Freeze margins from operational decisions before held-out evaluation; pilot variance determines replication, not acceptable degradation. Cluster uncertainty by source collection and task family. The data claim is supported only if the complete evidence-path metric meets its registered quality margin while governance error and valid-task refusal meet their respective bounds at the declared total cost. A result confined to injected policy faults cannot establish utility on natural tasks.

The P1 study remains a separate experiment. Adaptive learning, changed model routing, or persistent memory refinement must be frozen within a crossover period and reset across periods, or studied using parallel clusters. Record intended capability demand separately from the route actually executed; capacity shortages can change routing and otherwise masquerade as a capability intervention. The finite conformance artifact is preparation for those experiments, not a substitute for them.

\subsection{Agent-Side Memory: An Explicit/Implicit Dual Track}
\label{sec:memory-dualtrack}

The data substrate above is stack-external. A second memory surface lives inside the Harness: the agent-side record of its own execution, distinct from both the transient model context window and the contract-governed external substrate. We specify it as a dual track because the two halves have different owners, visibility, and change cadences.

\paragraph{Explicit memory (agent-visible).}
The explicit track records observable interactions along two organizational axes --- role (user, model, tool call, Skill call) and timeline (session, turn, step) --- plus a content type (input, intermediate artifact, log, output, error). Its indexing schema is the same 5W1H+Which record defined for the IR: \textit{What} is the content type, \textit{Who} the role together with producers, consumers, and explicitly excluded principals, \textit{When} the timestamps plus a validity window, \textit{Where} the source reference, \textit{Why} a declared rationale or causal hypothesis, \textit{How} the digest, key list, and short narrative, and \textit{Which} the temporal and cross-reference links. It records tool traces, declared checks, and evidence pointers rather than private chain-of-thought. Reusing one schema for agent memory and data-substrate facts is deliberate: it lets the asynchronous background indexing loop ingest agent traces as first-class facts, so that execution history composes with enterprise knowledge rather than living in a parallel silo.

\paragraph{Implicit memory (system-managed).}
The implicit track holds deployment configuration the agent never sees: tool and API call configuration, context-compression configuration, and storage-backend selection with partitioning and retention. The design rule is that changes to the implicit track are invisible to the agent, which consumes only explicit query results --- the memory-level instance of the control-plane/data-plane separation of Section~\ref{sec:manageability}.

\paragraph{Four resolved engineering decisions.}
Four design questions at this boundary have concrete resolutions, stated so they are falsifiable rather than aspirational.
\begin{enumerate}
\item \textit{Compaction offloads rather than silently deletes.} When the active context is compressed, retained original records move to governed archive storage; the near-line tier keeps derived summaries, indexes, and an offload pointer. Retention classes, legal holds, privacy limits, deletion requests, and access policy determine whether an original is archived, redacted, or deleted. Generating a final summary neither authorizes indefinite retention nor makes deletion safe.
\item \textit{Concurrent writes use a backend-neutral idempotency and ordering contract.} Multiple agents appending to the same memory submit an identity, scope, content digest, source version, and monotone sequence or logical timestamp. An implementation may serialize through a log, transactional store, or partitioned writer; no universal embedded single-writer design is assumed. Concurrency control for multi-agent systems is an active systems problem in its own right, with dedicated mechanisms proposed for it \cite{lyu2026coagent}; we cite that work to justify treating the ordering contract as a declared interface rather than to adopt any specific mechanism, and we make no throughput or contention claim of our own. Deduplication is valid only within the declared identity, scope, and version.
\item \textit{Capacity is a declared constraint with a hash-indexed overflow path.} Memory quota enters $\Omega$ alongside compute and token budgets. Near capacity, the coldest data is content-hashed and offloaded, leaving fixed-length index entries, so near-line volume scales with record count rather than content size --- and the same hash serves as the deduplication key of decision 2.
\item \textit{Compression and indexing are governed capabilities behind a stable interface.} One or more model-based or deterministic components may perform multimodal compression, merge, and index generation. Every derivative carries its producer and source versions. An upgrade requires benchmark regression on task-family fidelity, retrieval recall, evidence alignment, and join accuracy, plus revalidation of affected Skills whenever compatibility cannot be established from the declared contract. Version registration runs through Scaffold governance; semantic-contract ownership stays in the Harness.
\end{enumerate}

\paragraph{Cross-agent sharing.}
Agents do not read one another's memory directly. Access follows a governed three-stage path: ranked discovery over permitted metadata, structured navigation over 5W1H+Which and typed relations, then reading the authoritative record or evidence span. A natural-language summary may help rank or orient the search, but it is not the authority. Matching \textit{Who}, overlapping \textit{When}, and a typed \textit{Which} edge are policy and ranking features rather than guarantees of correctness. The sharing mechanism is specified, not measured; comparisons among lexical, embedding, graph/operator, and hybrid paths are registered as an open evaluation question with online latency and amortized indexing cost reported separately.

\section{Skill Lifecycle and Training}
\label{sec:skill-lifecycle}

\subsection{Skill-as-Code}
\label{sec:skill-as-code}

Treating a Skill as a prompt file makes capability growth difficult to review. Skill-as-Code applies a software lifecycle to the complete capability contract: source-controlled declarations, schema linting, effect review, contract tests, sandbox tests, model-version tests, signed release artifacts, staged rollout, monitoring, and rollback. That a Skill deserves the status of a software unit with its own identity, and that identity persistence across update, rollback, and removal is measurable rather than assumed, has been developed as an ontology in its own right \cite{fan2026skillware}; we take that framing as given. The broader case that routing agent behavior through executable, verifiable, stateful code is the direction of travel for harness design has likewise been surveyed \cite{ning2026codeharness}. What this subsection adds is narrower and specific to a probabilistic control plane: the point in the lifecycle at which a converged Skill stops being regenerated and becomes frozen code, and the release gate that decides when it may.

\begin{figure}[t]
\centering
\includegraphics[width=\linewidth]{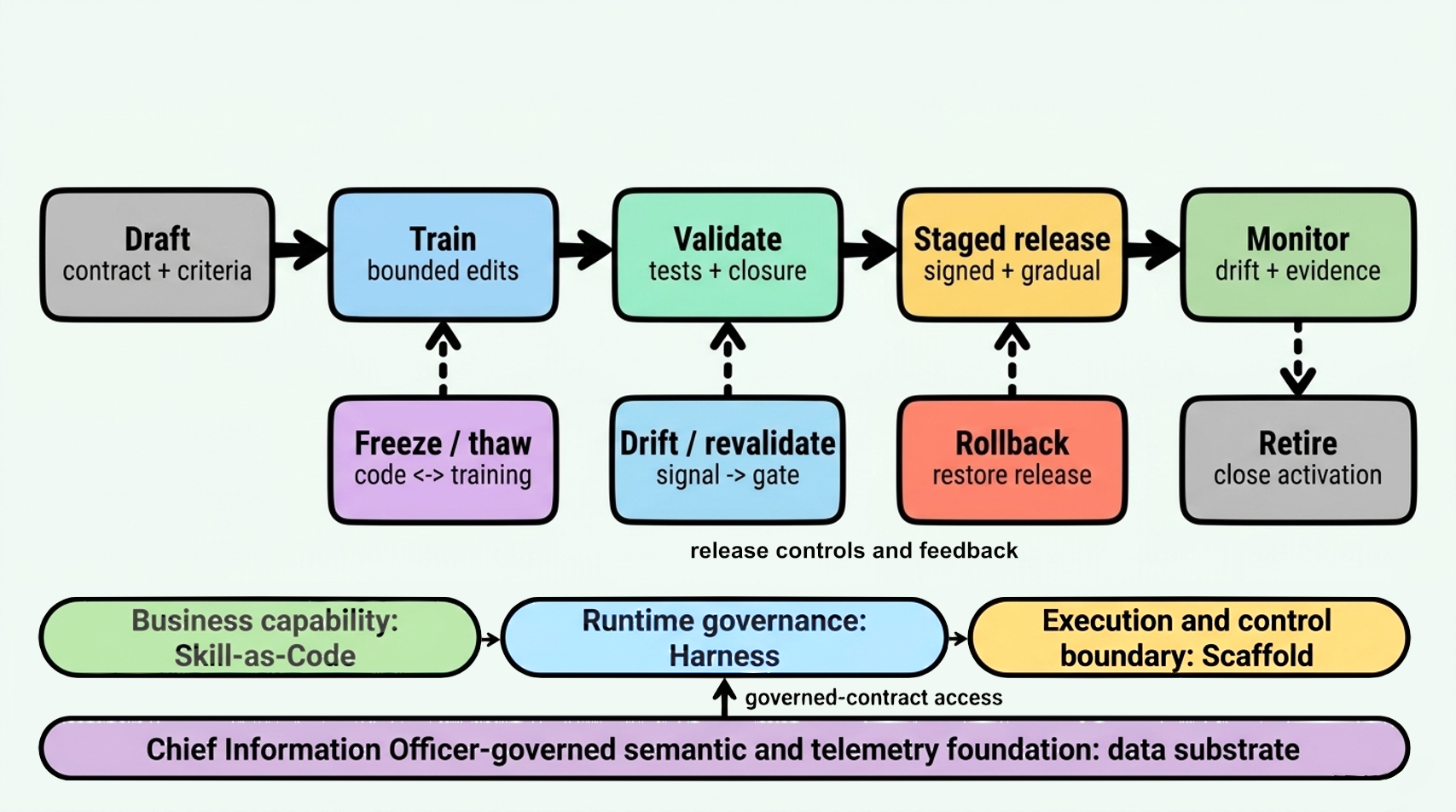}

\caption{Shown: a Skill moves through versioned validation, staged activation, monitored release, rollback, and retirement, with evidence attached to each transition and a freeze step where a converged Skill becomes code. Why it matters: the experiment can identify the capability configuration precisely and can repeat or reverse an intervention without conflating it with registry churn. Class: protocol.}
\label{fig:skill-as-code}
\end{figure}

Figure~\ref{fig:skill-as-code} makes validation and reversibility part of the release architecture rather than a documentation convention. The lifecycle is \textit{draft $\rightarrow$ train $\rightarrow$ validate $\rightarrow$ staged release $\rightarrow$ monitor}; observed drift returns the Skill to revalidation, a thaw returns frozen behavior to training, rollback restores a prior staged release, and retirement closes activation while preserving identity history. Operation closure is a validation gate: tests must show that externally visible behavior is expressible through declared effects. Model-version stability is tested at the same boundary, and a release is blocked if contract adherence, path selection, or postcondition satisfaction regresses.

Skill-as-Code also serves as a deterministic anchor in the probabilistic control plane. A Skill's operation space closes once its tool-set is stable, its output format is structured, and its decision branches are enumerable; at that point the execution process can be frozen into code rather than regenerated probabilistically each turn. Frozen code removes model sampling from that execution path and embeds a stable procedure, but its execution is deterministic only for fixed inputs, dependency and external-service snapshots, random seed, and execution semantics. Under those conditions, the control plane has probabilistic planning, deterministic frozen-code execution, and deterministic gates. Peilivanidis packages typed inputs, deterministic code and tests into versioned capabilities with model adapters \cite{peilivanidis2026equip}. Our proposed cross-version test fixes the bundle and inputs, records the loaded digest, and measures whether changing the planning model alters invocation or verified output. The frozen prefix of Section~\ref{sec:first-principle} stabilizes the context; the qualified frozen procedure stabilizes the decision within it.

\subsection{Training Before Freezing: A Dual-Subgoal Reward}
\label{sec:training-before-freezing}

Skill-as-Code explains how a stabilized Skill is governed and frozen, but not how the Skill reached stability. We assume a premise that this paper does not claim as its own contribution: the versioned Skill is trainable external state of an otherwise frozen agent, edited in text space under a \textit{bounded Skill-training loop} whose three elements are bounded edits, validation gates, and a rejected-edit buffer of refused proposals. Optimizer-level treatments of self-evolving Skills established that framing before this work; because the relevant sources predate the 2026-06-01 eligibility boundary of Section~\ref{sec:limitations}, they are recorded as design motivation rather than cited as evidence. What this subsection contributes is confined to the reward and admission structure of that loop. A single scalar is a lossy projection of a natural-language goal: it cannot tell the optimizer whether a failure lies in the outcome or in the process, nor which output sub-domain regressed.

We propose to structure the reward along two subgoal dimensions, both derived from artifacts the contract architecture already maintains. The \textit{outcome subgoal} scores the result separately along each sub-domain of the Skill's declared output schema $\sigma^{out}$, with an independent annotated benchmark per sub-domain and a weighted aggregate:
\begin{equation}
r_{\mathrm{out}}(s)=\sum_{d\in\mathrm{dom}(\sigma^{out})} w_d \cdot r_d\big(\mathrm{output}_d,\ \mathrm{benchmark}_d\big).
\end{equation}
The \textit{process subgoal} supplies a dense, immediate signal about observable execution: whether the correct tools and data sources were invoked, whether intermediate artifacts and declared checks satisfied the criteria, and whether each step gate passed:
\begin{equation}
r_{\mathrm{proc}}(s)=\frac{1}{|\mathrm{steps}|}\sum_k \mathbb{1}\big[\mathrm{step}_k \text{ satisfies the tool/data/logic/gate criteria}\big].
\end{equation}
Each mini-batch computes both subgoal families, and the validation gate becomes vector-valued rather than scalar. Let
\begin{equation}
R(s)=\left(\left(r_d(s)\right)_{d\in\mathrm{dom}(\sigma^{out})},r_{\mathrm{proc}}(s)\right).
\end{equation}
Before training, the evaluator fixes the verifier and tolerance $\epsilon_j$ for every component, together with protected components $P$ and target components $T$. A proposed Skill $s'$ is admitted relative to $s$ only when
\begin{equation}
\mathrm{Accept}(s'|s)=
\left[\bigwedge_{j\in P}R_j(s')\geq R_j(s)-\epsilon_j\right]
\land
\left[\bigvee_{j\in T}R_j(s')>R_j(s)+\epsilon_j\right].
\end{equation}
This Pareto-style gate prevents an aggregate gain from hiding a protected-domain regression while still requiring a material improvement on at least one declared target. The role of the human changes accordingly: instead of judging every iteration, the human confirms the evaluation method once, after which reward computation is automatic.

\begin{proposition}[Hypothesis P15: dual-subgoal gate benefit]
\label{prop:p15}
Compared with a scalar gate and with an output-vector-only gate, the output-plus-process vector gate $R(s)$ improves offline discrimination between Pareto-valid and dominated edits, improves attribution of a failure to a declared output sub-domain or process step, and improves online convergence at equal rollout budget without increasing protected-component regressions. The claim requires both comparator-specific contrasts to meet their preregistered margins.
\end{proposition}

\paragraph{Relation to concurrent work on regression control.}
The two decomposed-credit systems introduced in Section~\ref{sec:related} \cite{luo2026gsme,lin2026wml} reach the same conclusion that a single scalar is too coarse, and what remains distinct here is the axis along which credit is indexed. Both index it by an inferred property of the failure, discovered after the fact --- a pathology in one case, a workflow node and mechanism in the other. Our decomposition instead indexes credit by the sub-domains of the declared output schema $\sigma^{out}$, which exist in the contract before any run occurs. The practical consequence is that the reward dimensions are fixed by the same declaration that the Harness already gates and that Section~\ref{sec:derivation-closure} uses to compute the run-time satisfaction ratio, so training-time attribution and run-time completeness share one structure. Whether contract-derived dimensions outperform diagnosis-derived ones is an open question, not something we claim, and it is exactly the question posed by P15.

\paragraph{Caveat: trustworthiness of process criteria.}
One caveat constrains how the process dimension may be read. A deterministic testbed has shown that harness optimizers can propose guardrails for failure classes that provably never occurred, in 15 of 60 runs when the input merely resembled a familiar rule \cite{wang2026phantom}. We adopt that paper's term for the resulting artifact: a \textit{phantom violation} is a reported process failure that the recorded episode did not contain. Because $r_{\mathrm{proc}}$ depends on judging whether a step satisfied its criteria, a dimension-annotated rejection is only as trustworthy as the criterion that produced it. Process criteria must therefore be preregistered artifacts rather than model-generated at scoring time, and counterfactual no-violation episodes must test whether the gate invents a process failure.

As training converges, subgoals whose evaluation logic and process criteria stop changing have closed operation spaces; they cross the closure threshold of Section~\ref{sec:skill-as-code} and are frozen to code at the staged-release step. Training and compilation are two phases of the same Skill lifecycle.

\section{Six Measured Obligations}
\label{sec:obligations}

The six conditions below are not architectural virtues awarded by inspection. Each is a measured obligation with five required outputs: instrumentation coverage, observed violations, uncertainty, enforcement cost, and operating-region exclusions. For obligation $j$, an independent inventory supplies the eligible-event count $N_j$; the monitor supplies $O_j$ joinable event records and observes $X_j$ violations among them. Uncovered events are $M_j=N_j-O_j$. Self-reported Harness traces cannot establish their own completeness.

Each obligation declares a coverage floor $\gamma_j$, violation ceiling $\nu_j$, and monitor-sensitivity floor $\eta_j$ before collection. Coverage must satisfy $O_j/N_j\geq\gamma_j$. For support, every uncovered event is treated as a violation: the one-sided 95\% cluster-aware upper confidence bound $U_j$ for the worst-case rate $(X_j+M_j)/N_j$ must be below $\nu_j$ for every obligation. Bounds are not pooled across obligations. Because the decision rule also resolves the interaction-equivalence and semantic non-inferiority margins, the analysis plan must preregister a single family-wise error rule --- for example, joint coverage supplied by a simultaneous confidence region, or a closed testing sequence fixing the order in which obligations and margins are decided --- before collection; the per-obligation bounds are otherwise not jointly valid. If the denominator is unavailable or coverage falls below its floor, the verdict is inconclusive. With sufficient calibration and coverage, a lower violation bound above its ceiling gives conditional-engineering; a bound that overlaps the ceiling gives inconclusive. An upper bound below the ceiling passes this gate. These cases follow the precedence rule in Section~\ref{sec:decision-rule}.

Calibration is also mandatory. For every obligation, blinded diagnostic injections open preregistered known violation channels across the assigned cluster-periods. If $K_j$ violations are injected and $D_j$ are detected, the one-sided 95\% cluster-aware lower confidence bound for sensitivity $D_j/K_j$ must exceed $\eta_j$. Injected events are labelled and excluded from the experimental violation numerator and from semantic and runtime outcomes. Missing injections, a failed sensitivity floor, or a monitor that cannot be calibrated makes P1 inconclusive; high nominal coverage cannot substitute for calibration. The report includes all inventory, detection, injection, uncertainty, cost, and exclusion counts.

\paragraph{Typed closure.}
Instrumentation enumerates every node, input, output, tool, data source, authority, and external effect on each activated path and joins those events to the resolved contract. A violation is an executed or requested operation absent from the typed bundle, including a dynamically derived node that was not separately admitted. Coverage compares contract manifests with independent tool, network, and effector inventories. Uncertainty is reported for undeclared operations per activated path and per operation. Validation latency and manifest/evidence volume enter $E(c,s)$. Paths containing opaque executors whose operation inventory cannot be observed are excluded from $\Omega$ before outcomes are inspected.

\paragraph{Complete mediation.}
Instrumentation sits at every effectful boundary: tool dispatch, credential issuance, network egress, persistent write, message send, and privileged runtime transition. A violation is an invocation that reaches such a boundary without a current resolved-contract authorization. For executable calls, the join must include the digest observed at load or dispatch as well as the artifact selected at admission. A mismatch is a violation. An absent execution identity leaves the event uncovered. Coverage is the fraction of independently observed boundary events matched to a current admission record and its required execution evidence. The violation interval is cluster-aware because bypasses can share a process, node, or period. Interposition latency and controller capacity enter $E(c,s)$. Effectors that cannot expose an independent call inventory are pre-outcome exclusions.

\paragraph{Effect non-interference.}
Instrumentation labels writes, messages, cache mutations, model-side state, and externally visible effects with contract and tenant identities, then records readers and downstream consumers. A violation is an undeclared path by which one admitted run changes the authorized effects or semantic inputs of another. Coverage includes every mutable or communicative channel in the threat model, not only database writes. Uncertainty is reported by channel and cluster-period. Labeling, taint or provenance processing, and isolation cost enter $E(c,s)$. Shared services without attributable mutation or consumption logs are excluded from $\Omega$.

\paragraph{Shared-state isolation.}
Instrumentation records every read, write, lock, cache access, session update, and durable checkpoint against an assigned state partition. A violation is an access outside the contract's partition or an undeclared contention edge between treatment units. Undeclared contention is not hypothetical: concurrent multi-agent execution has motivated dedicated concurrency-control designs \cite{lyu2026coagent}, which is why this obligation is instrumented rather than argued. Coverage uses storage, cache, lock-manager, and process-level inventories. The report gives interval estimates for unauthorized accesses and cross-partition contention. Namespace, copy, flush, and reset costs enter $E(c,s)$. State that cannot be reset, snapshotted, partitioned, or audited is excluded before period assignment.

\paragraph{Resource invariance.}
Instrumentation hashes the semantic inputs visible to the activated behavior, including model settings, token limits, tool schemas, policy facts, data snapshot, clocks where relevant, and resource metadata exposed to the model or tools. A violation occurs when changing compatible $s$ changes an undeclared semantic input or crosses a preregistered resource boundary that defines a different capability environment. Coverage is the fraction of semantic-input fields independently captured in every arm. Field-level differences and their uncertainty are reported, not averaged into one score. Capture and normalization costs enter $E(c,s)$. Accelerator classes, memory limits, or degradation modes known to alter semantics are separate strata or outside $\Omega$.

\paragraph{Scheduler independence.}
Instrumentation records queue state, priorities, admission feedback, retry decisions, autoscaling signals, placement hints, cancellation, and scheduler-written metadata visible to the Harness, model, tools, or data path. A violation is an undeclared scheduler-to-semantics or capability-to-placement feedback edge. Coverage joins scheduler decisions to every admitted and rejected run. Violation intervals are estimated by cluster-period and failure regime. Scheduler logging, policy evaluation, and any traffic-shaping overhead enter $E(c,s)$. Adaptive schedulers whose policy version or feedback channels cannot be pinned and observed are excluded from $\Omega$.

Passing all six thresholds does not prove that no seventh coupling channel exists. It establishes that the preregistered threat model was measured well enough to apply P1 within $\Omega$. The required diagnostic injections test the monitor assigned to each declared channel; they do not expand the claim beyond that threat model.

\section{Cluster-Period Crossover Study}
\label{sec:evaluation}

\subsection{Unit, Assignment, and Interventions}

The experimental unit is a cluster-period, also called a system epoch: an isolated worker pool, tenant partition, or deployment cell operated under one assigned $(c,s)$ pair for a fixed period. Requests inside a period are repeated observations, not independent randomized units. Cluster-period assignment avoids pretending that requests sharing caches, queues, schedulers, or controllers are independent. This is the paper's only unit of randomization; earlier drafts that randomized individual runs are superseded.

Clusters cross over through all eligible capability and Scaffold configurations. Period assignment is randomized within prespecified cluster and failure-regime blocks. Order balancing uses counterbalanced sequences so each treatment appears comparably early and late and follows every other treatment often enough to estimate differential carryover. The design follows established crossover and cluster-crossover principles \cite{jones2014crossover,turner2007cluster}. Period and sequence terms are fixed before outcome inspection, but neither term substitutes for an explicit lagged-treatment carryover estimate.

A capability intervention changes $c$ by activating a versioned bundle that occurs on admitted paths in the fixed workload. Adding inactive registry entries is not a treatment. A capacity intervention changes $s$ by altering compatible worker count, resource class, or topology while logical admission policy, activated bundle, model settings, workload mix, and data snapshots remain fixed. Both interventions must be observable in resolved contracts and physical resource records.

Two confounds require explicit control on the capability axis. First, retrieval competence must be fixed and reported, because a controlled study of progressive disclosure found that its benefit shrinks toward zero when the harness already partitions and retrieves competently, and becomes decisive only at corpus scale \cite{he2026disclosure}; run both a single-source and a many-source scale. Second, hold activation depth at one level unless depth is the preregistered intervention, as the same work reports that a second routing layer never helped.

\subsection{Reset, Workload, and Failure Regimes}

Every transition uses a prespecified reset or washout procedure. It drains or cancels in-flight work, restores mutable data and policy snapshots, clears or versions caches and sessions, resets rate-limit and retry state, reinitializes scheduler history, and waits through a measured stabilization interval. Millisecond-level sandbox checkpoint and rollback is one mechanism by which such a reset could be made affordable \cite{dong2026deltabox}; whether it suffices for the sentinels below is untested.

The predeclared reset sentinel metrics are: zero prior-period contract identities in queue, cache, session, retry, rate-limit, and scheduler-state inventories; exact data, policy, model, tool, and assigned-treatment manifest hashes; a bounded 90\% confidence interval for sentinel log-p95 latency relative to a fresh-start reference; and a bounded one-sided 95\% lower confidence bound for the sentinel every-requirement satisfaction risk difference. All conditions must pass before the next period begins. The two numeric bounds are deliberately left unfixed here: under the margin rule of Section~\ref{sec:p1} they must be derived from a named decision with a recorded owner, and the placeholder values used in earlier drafts ($\pm\log(1.05)$ and $-0.025$) carried no such derivation. A study that registers this protocol must supply them.

A sequence is primary-analysis eligible only if every assigned period completes and every transition passes all reset sentinels before outcomes are unmasked. A failed transition invalidates the complete sequence; partial sequences do not enter the primary estimator. The power plan fixes a minimum number $K_{\min}$ of complete sequences per treatment order. Loss of order balance or fewer than $K_{\min}$ eligible sequences for any order makes P1 inconclusive. Reset-failure and sequence-completion rates are preregistered and reported by incoming treatment, preceding treatment, period, and order before outcome unmasking. The complete-sequence estimator is paired with an assignment-based intention-to-treat sensitivity analysis over every randomized cluster-period: reset failure counts as an operational failure for $Q_{\mathrm{req}}$, missing $R$ is bounded by prespecified worst-case values, and a tipping-point analysis varies outcomes for failed transitions. Treatment imbalance in reset failure, or a P1 decision that changes within the sensitivity range, makes P1 inconclusive.

The workload mix, arrival process, task corpus, identity distribution, model and tokenizer, tool versions, external-service emulator or contract, and data snapshots are fixed across crossover arms. Every eligible cell is repeated over preregistered sampling seeds, arrival-process seeds, and system seeds. Failure regimes are crossed or blocked explicitly: normal operation, worker loss, delayed external service, quota pressure, and one or more injected recoupling channels. The base P1 decision and diagnostic failure-regime results are reported separately.

Pre-outcome exclusions are determined from admission, reset, instrumentation, and infrastructure-health facts before semantic or runtime outcomes are revealed. Permitted exclusions include failed reset sentinels, corrupted treatment assignment, unavailable independent instrumentation denominator, a noncompatible Scaffold class, or an external incident that violates $\Omega$. Load-induced failure is an outcome, not an exclusion. Exclusion counts and cluster-period identities remain in the report.

\subsection{Estimands and Uncertainty}

The primary $R(c,s)$ analysis estimates $\Delta_R$ on the specified log-p95 scale with cluster and period structure, sequence and failure-regime terms, and prespecified workload strata. The model includes first-order lagged capability, lagged Scaffold capacity, their lagged interaction, and their preregistered interactions with current treatment. First-period lag terms are undefined and do not contribute to carryover contrasts. Differential carryover is acceptable only when every 90\% confidence interval for a lag-by-current-treatment contrast on $Y_R$ lies inside $[-m_R/2,m_R/2]$ and every one-sided 95\% lower bound for the corresponding $Q_{\mathrm{req}}$ risk difference exceeds $-m_Q/2$. An unresolved or out-of-bound lag effect makes P1 inconclusive even after reset sentinels pass.

Cluster-aware uncertainty uses a prespecified mixed-effects model with a small-sample correction, checked by cluster-level randomization inference when the order design permits \cite{turner2007cluster}. The exact covariance structure, finite-sample correction, multiplicity adjustment for more than two treatment levels, and fallback when too few clusters remain are fixed in the analysis plan. Request-level standard errors are not valid substitutes. The $Q(c,s)$ analysis uses the $Q_{\mathrm{req}}$ risk differences and non-inferiority rule of Section~\ref{sec:p1}; aggregate task scores remain secondary so that one strong subtask cannot hide a required failure. Aggregate recall is insufficient on its own: a white-box study observed requirement coverage above 92\% while task outcomes collapsed, because a small number of dropped requirements can invalidate an otherwise complete artifact \cite{xue2026longcontext}. The report therefore gives the fraction of runs in which every declared requirement was satisfied and records which requirements were dropped rather than only how many.

\paragraph{Enforcement overhead is a marginal quantity.}
The $E(c,s)$ counterfactual is one baseline: paired safe replay in a sealed emulator. For every sampled immutable request or event trace in each $(c,s)$ cell, the emulator runs an enforcement-on replay and a minimal pass-through replay with identical capability and Scaffold configuration, seeds, snapshots, model and tool outputs, and safe stub effectors. The minimal pass-through arm consumes the prevalidated trace but omits the admission, mediation, isolation, and evidence operations whose runtime cost is being estimated. Replay order is randomized within each pair. Out-of-band emulator containment remains active in both arms and is excluded from runtime enforcement cost. For component $k$, the cell estimator is the mean paired on-minus-baseline difference $\widehat{E}_k(c,s)$.

Because both replay arms are pinned to identical model and tool outputs, $E(c,s)$ estimates the \textit{marginal runtime enforcement overhead} only. It deliberately excludes the opportunity cost of rejected behavior: when admission refuses an operation, the model's subsequent trajectory changes, and that downstream consequence is fixed away by the pinned outputs. That cost is not lost from the accounting --- it appears in the semantic outcome $Q(c,s)$, where a refused-but-necessary operation shows up as an unsatisfied requirement. Reporting $E$ without this statement would invite the reading that enforcement cost has been underestimated; the two quantities are complementary and must be read together.

The analysis reports admission, mediation, isolation, reset, evidence-processing, and total components on their preregistered natural-unit scales. Every one-sided 95\% upper confidence bound must lie below its component budget $B_{E,k}$ in every required $(c,s)$ cell. A missing pair or unresolved bound is inconclusive; a lower confidence bound above budget falsifies P1.

\paragraph{Power.}
Power is designed around the interaction-equivalence and semantic non-inferiority margins, not the larger and easier Scaffold main effect. Detecting an interaction contrast of a given magnitude requires substantially more replication than detecting a main effect of the same magnitude, because the contrast is a difference of differences and accumulates the variance of every cell entering it; a design powered only for the main effects will routinely fail to resolve the term P1 actually rests on. The plan states clusters, periods, observations per period, assumed intracluster correlation, carryover allowance, minimum detectable interaction, semantic non-inferiority margins, and expected attrition from pre-outcome exclusions.

A mean-based design-effect formula is not an adequate power calculation for a period-level p95 endpoint. Quantile uncertainty depends on the latency distribution near the percentile, timeout mass, within-period serial dependence, and cluster-period heterogeneity. Before the confirmatory study, simulate complete randomized sequences under a pilot-calibrated data-generating model, retaining queueing dependence, censoring, carryover, and reset attrition. Apply the exact planned estimator and verdict rule to every simulated dataset and choose replication for the registered equivalence and non-inferiority margins. Publish sensitivity curves over plausible tail shapes and correlations. Pilot data may determine nuisance parameters and replication, but cannot widen the decision-derived margins. No minimum cluster count or achieved power is asserted in this paper.

\begin{figure}[t]
\centering
\renewcommand{\arraystretch}{1.25}
\small
\begin{tabular}{p{0.15\linewidth}p{0.26\linewidth}p{0.22\linewidth}p{0.22\linewidth}}
\hline
\textbf{Hypothesis} & \textbf{Intervention and control} & \textbf{Evidence plane} & \textbf{Falsification or inconclusive condition} \\
\hline
\textbf{P1} (primary): capability $\times$ Scaffold separability &
Cluster-period randomized crossover. Independently vary activated capability configuration $c$ and Scaffold capacity configuration $s$; hold workload and data snapshot fixed; preregister reset or washout. &
Semantic outcome $Q(c,s)$; runtime response $R(c,s)$; enforcement overhead $E(c,s)$; six condition-violation rates; cluster-aware uncertainty; capability-by-Scaffold interaction estimand. &
Insufficient evidence: inconclusive. Clearly violated obligations: conditional-engineering. Only after all obligation gates pass, resolve runtime interaction, semantic non-inferiority, and cost bounds as supported, falsified, or inconclusive. \\
\hline
P15: vector gate & proposal-bank contrasts & -- & margin failure or phantom violation \\
P16: evidence-path sufficiency & held-out joint contrast & -- & direct-reading control wins \\
P17: policy-contract decoupling & dependency-change oracle & -- & unanticipated propagation \\
\hline
\end{tabular}

\caption{Shown: P1 maps to cluster-period interventions on capability and capacity, six measured obligations, semantic and runtime estimands, enforcement cost, and the conditions distinguishing supported, falsified, conditional-engineering, and inconclusive verdicts. Why it matters: the matrix ties every architectural claim to an observation and prevents missing coverage or low power from being reported as separability. Class: proposed measurement design.}
\label{fig:evaluation-matrix}
\end{figure}

\subsection{Reporting}

Figure~\ref{fig:evaluation-matrix} summarizes the program: the interventions, the six obligations, the three estimands, and the conditions under which each of the four verdicts is reached. Its purpose is to make missing coverage and insufficient power visible as named outcomes rather than letting either be reported as separability.

The primary result table reports each $(c,s)$ cell, cluster-period count, exclusion count, six coverage rates, six violation intervals, $R(c,s)$, $Q(c,s)$, $E(c,s)$, the interaction estimate and interval, all preregistered margins, and the final four-way decision. Capacity-response curves remain visible even when P1 is inconclusive. This preserves the useful positive Scaffold main effect while refusing to infer independence from it.

\subsection{A Minimal Viable First Study}
\label{sec:eval-minimal}

The protocol above describes the program at full extent, which no single group is likely to execute first. Because the largest risk to this line of work is that its evidence requirements are heavy enough to postpone all evidence indefinitely, we state a small configuration that can produce inspectable engineering evidence relevant to P1.

The minimal study declares a deliberately small $\Omega$: a single tenant, one workload family, one model and tokenizer version, one policy snapshot, and exactly two compatible Scaffold classes rather than a topology sweep. The capability axis uses synthetic Skills whose declared inputs, outputs, and effects are authored for the experiment, so that ground truth about declared surfaces is available without waiting for a production Skill population. The factorial is correspondingly small --- a handful of capability configurations crossed with two capacity settings --- and replication is spent on cluster-periods per cell rather than on additional cells.

What this configuration must not reduce is instrumentation. All six obligations are instrumented and all six violation rates reported, because the rates are the study's primary product even when the outcome contrasts are underpowered. Instrumenting an obligation is not the same as passing it, and the minimal study does not require that all six pass: a first study whose complete-mediation violation lower bound exceeds its ceiling while the other five obligations pass yields a conditional-engineering result that names exactly which channel a real runtime leaves open and what it costs to close, and that result is more useful to the next study than an unreported failure to reach the conjunction. Two economies are acceptable at this scale: the diagnostic injection arms may be reduced to the shared-compute and external-service quota channels, and the semantic estimand may be restricted to task success, every-requirement satisfaction, and the five interface semantics.

The expected output is modest and specific: six violation rates with the instrumentation that produced them, a semantic interval across two Scaffold classes, a two-point capacity response, an interaction estimate reported with its achieved power, and a cost account of what closing each violated obligation required. A study of that size cannot establish separability and should not claim to. It can establish whether the six obligations are instrumentable at all in a running system, which is currently unknown and is a precondition for every larger study.

\section{Secondary Protocols}
\label{sec:secondary}

The architecture suggests narrower studies that do not enter the P1 decision unless preregistered as part of an obligation. Each is stated with its refuting condition.

\paragraph{Harness-bypass ablation.}
Execute the same workloads with full mediation and with narrowly instrumented bypasses: direct tool invocation, untyped data fetch, unrecorded retry, and direct Scaffold selection. Count undeclared effects, policy violations, evidence omissions, non-replayable decisions, incident localization time, and false rejections. The mediation claim is weakened if an unbound call reaches an effector without a typed rejection in the trace, or if bypass does not measurably affect control or replay under adversarial workloads.

\paragraph{Control-plane leakage.}
Seed the registry with canary metadata, deprecated schemas, and policy text the workload never requires. Compare eager loading, retrieved activation, and opaque-reference activation; track canary reproduction, prompt tokens, deprecated-tool selection, policy-version confusion, and plan accuracy. The claim fails if control metadata appears in outputs or influences plans despite opaque activation, or if opacity prevents the planner from satisfying valid requests.

\paragraph{Path-level composition safety.}
Construct Skill pairs and longer paths that are benign individually but violate data-flow, privilege, or effect-ordering constraints when composed \cite{xie2026composition}. Compare local allowlists with path-aware admission; measure hazardous-path detection, false positives, dynamic-graph coverage, and decision latency. The claim is weakened if invariants cannot express realistic hazards, if dynamic insertion routinely escapes checks, or if false positives make valid composition impractical.

\paragraph{Inline expansion versus bounded derivation.}
Contrast granting authority inline with the bounded derivation of Section~\ref{sec:derivation-closure}. The derivation protocol is supported only if inline expansion measurably widens the activated authority surface or the trace.

\paragraph{Dry-run and locality economics.}
Hold graph and Scaffold pool fixed; compare immediate execution with dry-run admission, and locality-oblivious with locality-aware binding. Report dry-run latency, avoided failed-run cost, bytes moved across zones, completion latency, resource utilization, and policy rejection stage. Dry-run is not justified where its predictions are poorly calibrated or its cost exceeds avoided work.

\paragraph{Skill-as-Code across model versions.}
Freeze Skill source and test corpus while varying planner and executor model versions; include perturbations of input phrasing, tool errors, and partial data. Separate the component under test from the component that scores it, and grade the first attempt only: without that separation an agent can silently repair itself mid-run and have the repaired output scored as a pass, inflating pass rates toward a spurious ceiling \cite{anand2026aeval}. Log every self-correction as an observation rather than crediting it. The lifecycle is insufficient if production-significant changes pass the suite, or if nondeterminism makes the suite too unstable to gate releases.

\paragraph{Dual-subgoal reward for Skill training (P15).}
Construct one fixed proposal bank for a multi-sub-domain task and evaluate every proposal under three gates: (a) one scalar; (b) the output vector; and (c) the output-plus-process vector $R(s)$. Preregister all verifiers, tolerances, protected and target sets, process criteria, decision thresholds, and the two comparator-specific contrasts $(c)-(a)$ and $(c)-(b)$; both must meet their margins, and averaging the comparators is insufficient. Independent experts or deterministic tests assign blinded oracle labels for Pareto validity, protected-component regression, and process-attribution truth. Include counterfactual episodes with no violation of a selected process criterion as a phantom-violation oracle \cite{wang2026phantom}. Then run the bounded training loop with each gate under matched multi-seed conditions, holding model, workload distribution, optimizer prompt, rollout budget, verifier versions, and stopping rules fixed. P15 is weakened if configuration (c) fails a preregistered discrimination, attribution, convergence, or rollout-cost margin against either comparator, increases protected-component regressions, or reports process violations that the phantom-violation oracle disproves.

\paragraph{Evidence-path sufficiency and policy-contract decoupling (P16, P17).}
For P16, preregister the task family, fields, output descriptor, source snapshots, access budget, reconstruction metric, and acceptance margins before splitting reports into strict training, validation, and held-out sets. Build and revise the address/index policy using training and validation only, then pin its sources, operators, fallback rules, and verifier before inspecting the held-out set once. On identical held-out tasks, randomize the agent to: (a) the pinned address/index policy with governed source fallback; (b) token-matched direct reading; (c) metadata-only input; (d) preregistered field or operator ablations; or (e) shuffled indexes assigned to the wrong source. Task-family sufficiency requires policy (a) to be equivalent or non-inferior to (b) while exceeding the metadata-only, shuffled-index, and informative ablation controls by their positive margins. P16 is weakened when that joint contrast fails, when source or evidence identities cannot be reconstructed, or when leakage auditing finds unauthorized access, held-out-guided revision, or an unpinned verifier.

For P17, build a versioned Data Wiki, Theme Wiki, and IR over an enterprise-like corpus. Before each intervention, preregister compatibility ranges and an expected dependency graph for three change classes in each direction: a compatible entry-content update; an incompatible semantic or evidence-contract change requiring migration; and a cross-registry use-case change expected to update the IR and named entries. A named \textit{artifact-change oracle} records touched entries, schemas or modules, IR records, tests, generated artifacts, and hidden-state dependencies, then compares the observed change set with the preregistered graph while reviewers label extra and missing propagation blinded to the claimed decoupling. P17 is weakened when a declared-compatible one-sided change causes unanticipated propagation into the other registry, or when an undeclared shared dependency bypasses the IR; expected migrations do not by themselves falsify it.

\paragraph{External-data reconstruction.}
Test whether source, snapshot, and access-decision identity can be rebuilt from the returned evidence bundle of Section~\ref{sec:data-substrate}. It fails if an evidence bundle cannot reproduce those identities.

Automated Skill training beyond P15, cross-registry change propagation, prefix-stability measurement, and seed-affinity placement remain future work. They require their own interventions, oracles, and evidence boundaries, and should not be attached to P1 as additional propositions, because doing so would turn one systems question into an open-ended architecture survey.

\section{Discussion}
\label{sec:discussion}

\subsection{Enterprise Adoption as Contract Adoption}

The proposed architecture does not require four new departments. It requires four responsibility objects to remain distinguishable even when one organization owns several of them. A business product owner can change a Skill without acquiring authority over runtime policy; a Harness team can change admission and composition without silently redefining business meaning; a Scaffold team can evolve capacity while remaining accountable for its NFRs; and the CIO-governed data substrate can evolve semantic and telemetry contracts on its own cadence. This separation gives architecture reviews a concrete agenda: owner, version, stable contract, evidence obligation, rollback authority, and permitted rate of change.

Adoption should be evaluated on both planes in Table~\ref{tab:evidence-planes}. Neither success plane substitutes for the other, and a release process that reports only one leaves a different enterprise decision unanswered.

The Harness also creates a control-plane concentration risk. If its compiler, registry, or policy evaluator is unavailable, admission may stop; if one is compromised, a typed contract can become false assurance. Implementations need replicated control services, signed artifacts, independent attestation, and a declared degraded mode. New requests should fail closed when compilation is unavailable. Cached decisions require explicit freshness bounds, and registry failure may preserve already activated versions for in-flight work without admitting new ones. Degradation should reduce capability rather than widen governance.

Harness mediation and stack-external data access also have costs. Contract compilation, invariant evaluation, telemetry integration, semantic joins, and evidence retention consume latency, capacity, and administrative effort. Their cost must be measured against avoided failures and improved auditability; this paper provides no measured overhead. Selective activation can miss a relevant Skill, strict closure can reject useful exploration, path checking can be expensive for dynamic graphs, trace retention can conflict with privacy, and physical decoupling can fail under specialized hardware or residency constraints. The contracts make these trade-offs attributable; they do not remove them.

\subsection{Dispute and Escalation}

The organizational thesis is testable only if the contracts also state how contested decisions are resolved, so we fix four rules. First, an ownership dispute over a Skill is resolved against the stable contract of record: the declared owner of the versioned artifact governs until a named escalation authority changes the record, and the change itself carries the authority's identity, the evidence considered, and a timestamp in the trace. Second, a contested release is held in staged state until the named authority confirms it against the evidence planes of Table~\ref{tab:evidence-planes}; one plane's evidence alone cannot push a release. Third, if the data substrate is unavailable or a governing authority cannot be reached, the system degrades by reducing capability rather than widening governance. Fourth, every escalation is recorded as part of the evidence and audit trail, so the dispute-resolution path is itself auditable and attributable rather than a silent override.

\subsection{Artifact and Preregistration Statement}

The ancillary directories \texttt{artifact\_v29/} and \texttt{scaffold\_v30/} contain the data-use and physical-binding fixtures, their controls, and their results. Sections~\ref{sec:ir-conformance} and~\ref{sec:scaffold-binding} define the executed domains. There is no completed runtime scaling or retrieval-quality experiment. The remaining protocols require workload-specific decisions, numerical tolerances, pilot nuisance parameters, and preregistration before execution. They are protocol templates, not a completed preregistration, and no claim of ready-to-run confirmatory replication is made.

Three commitments follow for any study that this paper's authors later conduct against these protocols. First, the operating region $\Omega$, the six obligation instrumentations and their thresholds, all equivalence, non-inferiority, and interaction margins, the replication per cell, and the resulting minimum detectable interaction will be registered before data collection and published with the results whether or not the outcome favors the conjecture. Second, the obligation decisions will be frozen before any outcome estimand is unblinded, and the observed violation rates will be reported alongside the binary decisions. Third, a study that ends in the conditional-engineering state will be reported as such rather than withheld, since its violation rates and closure costs are the information the next study needs.

\subsection{Open Adoption and Research Questions}

Enterprise adoption raises organizational and technical questions together. What is the smallest contract language that captures effects without becoming a second programming language? Which decision rights and change cadences work when one team spans several responsibility objects? Which business metrics should block a Skill release, and which Scaffold NFR margins should block runtime admission? How should business and runtime telemetry be joined without transferring data ownership into the Harness? Which graph invariants can be checked statically, and which require runtime monitors? Which diagnostic injections localize recoupling, and at what workload complexity do dry-run and semantic-join costs become economical? A further question is comparative: no protocol here contrasts this contract boundary with the orchestration frameworks currently in production use, and a per-dimension comparison against them remains the most conspicuous gap in the evaluation program.

\section{Limitations and Threats to Validity}
\label{sec:limitations}

This paper presents a reference architecture, two finite synthetic contract-conformance studies, and proposed confirmatory experiments. The binding fixture assumes trustworthy measurements and an authoritative status service. A deployment must independently test fencing of stale workers, atomic deduplication across replicas and the interval between a successful check and an effect. It is not a production system or enterprise adoption study. It provides no completed runtime implementation, no measured enforcement budget, no natural-task evaluation dataset, no runtime benchmark result, and no claim about a deployment's supported scale. It does not demonstrate operation from 0 to 100{,}000 admitted or active agents. It does not establish business value, semantic-join quality, telemetry completeness, portability, safety, or any Scaffold NFR. The diagrams are responsibility and protocol views, not screenshots or implementation evidence.

A verdict of \textit{supported} would not mean that independence has been proven. It would mean only that within the preregistered $\Omega$, margins, power, and instrumentation, no falsification criterion was triggered. Any stronger extrapolation is unwarranted.

The responsibility mapping may underrepresent human approvals, long-running state, multi-tenant incentives, cross-organization trust, and regulated decision rights. Skill, Harness, and Scaffold are overloaded terms, and existing systems may combine their responsibilities in one service. The organizational and governance side of the thesis remains the least evidenced part of this work: owner division, dispute escalation, and cross-enterprise transferability have not been examined by any organizational study.

The evidence base is layered rather than uniform. Classical foundations are stable anchor points, while the retained contemporary sources were released in the months preceding this revision. They motivate governance, composition, lifecycle, regression control, structured change, and the harness-scaling thesis itself, but they do not validate this architecture or its organizational claim. Fast-moving preprints may also change after the versions recorded in the bibliography.

Separability may hold only within a bounded operating region. A new capability can require a new hardware class; a new Scaffold can expose timing or locality behavior that changes task quality. The six obligations may be incomplete, correlated, or difficult to hold fixed, while deterministic gates and trace identity can expose a violation without preventing it. Strict coverage rules reduce false support but can make useful deployments inconclusive or force a conditional-engineering verdict. How often the full conjunction is reachable in a production runtime, and at what engineering cost, is unknown to us; it is possible that the conjunction is expensive enough that P1 is testable only in a reduced setting whose external validity is then in question.

Equivalence, non-inferiority, and interaction conclusions depend on chosen margins and achieved power. Margins can be chosen too loosely, and enforcement budgets can omit organizational or operational cost. The crossover improves efficiency when clusters can return to a comparable state; it is unsuitable for irreversible learning, persistent user adaptation, or infrastructure changes with long carryover, which need parallel clusters or longer stepped designs. P15, P16, and P17 remain hypotheses or propositions with proposed controls, not observed results. The finite synthetic conformance result in Section~\ref{sec:ir-conformance} tests a narrower reference model and does not validate any of those hypotheses.

The formal data appendix has a narrower scope than the architecture. It assumes finite alphabets, deterministic query-agnostic encoders, exact task answers, and separate accounting for persistent representation and online communication. Randomized encoders, approximate or continuous tasks, learned-retrieval calibration, changing source distributions, and total lifecycle cost require additional analysis. Its impossibility statements apply only when an interface merges source states that the declared task family must distinguish; they do not establish that one retrieval technology is universally superior or that retaining a source by itself makes the source accessible, correct, fresh, authorized, or auditable.

\section{Conclusion}
\label{sec:conclusion}

Enterprise AI needs a contract across groups that change different things at different rates. This paper assigns reusable, versioned business capability to Skill; runtime compilation, admission, composition, governance, and evidence to Harness; the execution/control boundary and NFR ownership to Scaffold; and semantic joins plus integrated telemetry to a stack-external, independently governed data substrate. Architecture as a shared organizational contract is the paper's narrower organizational thesis, not a claim that one organization chart fits every enterprise.

The scientific core is cost-aware capability-capacity separability. P1 asks whether activated independently deployable behavior on admitted paths preserves the capacity-response relationship, whether compatible capacity changes preserve semantic outcomes, and whether the required controls fit an enforcement budget. A cluster-period randomized crossover, six measured obligations, explicit decision-derived margins, and a four-state verdict make that question answerable. Deterministic gates and trace identity make departures detectable, auditable, and attributable; they do not cause independence.

The paper therefore asks enterprises to demand two forms of proof. Business/use-case evidence must show that a Skill produces the intended workflow outcome. System/runtime evidence must show that the admitted path and its execution satisfy contract and NFR obligations. Neither substitutes for the other. The two executable fixtures make different errors observable: reading a source after its authority has changed and substituting a worker whose actual execution profile differs from the one admitted. Their finite controls provide concrete failure cases for calibrating the deployment monitors. The next scientific steps are the held-out data study of Section~\ref{sec:data-study} and the minimal runtime study of Section~\ref{sec:eval-minimal}: implement the monitors, run one small crossover, and publish the result including the cases that end as falsified, conditional-engineering, or inconclusive.

\section{Technical Appendix: Information-Theoretic and Discrete Foundations}
\label{sec:formal-data-foundations}

This appendix formalizes one narrow question raised by the external data substrate: what can a representation fixed at write time preserve when the eventual query is not yet known? The results justify retaining governed source evidence and treating indexes, summaries, and relations as task-bounded derivatives. They do not prove the empirical superiority of the proposed Data Wiki, and they do not remove storage, governance, or retrieval costs.

\subsection{Scope, Notation, and Deficiency}

Let $X$ be a source record on a finite alphabet, $Q$ a query from a finite declared task family, and $Y=y(X,Q)$ the exact answer. For the partition statements, every source-query pair in the declared Cartesian domain has positive joint probability; without this support condition, zero conditional entropy gives only almost-sure sufficiency on the observed joint support. A deterministic write-time encoder produces $Z=f(X)$ before the realized query is available. A read-time decoder produces $\widehat{Y}=g(Z,Q)$. Entropy and the conditional Fano step use standard information-theoretic definitions \cite{cover2006elements}. The query-conditional information deficiency of the representation is
\begin{equation}
\Delta(f)=H(Y\mid Z,Q).
\end{equation}
The encoder is exactly sufficient for the declared task family when $\Delta(f)=0$, equivalently when some decoder recovers $Y$ for every source-query pair in scope. All conclusions are relative to that task family and support. A randomized encoder, a query-conditioned materialization, an approximate answer relation, or a continuous source requires a different sufficiency notion and is outside the propositions below.

\begin{proposition}[Decoder-independent error floor for a fixed representation]
Let $M\geq2$ be the answer-alphabet size and let $P_e$ be the error probability of any decoder using only $(Z,Q)$. Then $\Delta(f)\leq H_b(P_e)+P_e log_2(M-1)$. Consequently, when $\Delta(f)>0$, every decoder has a positive inverse-Fano error lower bound determined by the fixed representation and task. A stronger agent may reduce decoder suboptimality, but it cannot reconstruct distinctions absent from $(Z,Q)$. Proof: apply the conditional Fano inequality given $Q$ and average over $Q$. The inequality supplies a lower bound, not a point estimate of observed error, and equality need not hold.
\end{proposition}

\subsection{Partitions and the Minimal Sufficient Representation}

Every deterministic encoder induces a partition $\pi_f$ of the source alphabet: two source states share a block exactly when the encoder maps them to the same value. Each query $q$ induces a task partition $\pi_q$ whose blocks contain source states with the same answer to that query. Let $\pi^\star$ be the common refinement of all $\pi_q$ in the declared task family; it is the coarsest partition that still preserves every task-relevant distinction.

\begin{proposition}[Partition criterion, rate bound, and query monotonicity]
The encoder is exactly sufficient if and only if every block of $\pi_f$ is contained in a block of $\pi^\star$. Therefore every sufficient deterministic representation satisfies $H(Z)\geq H(\pi^\star)$, and the block identifier of $\pi^\star$ achieves equality, so $R_{\min}=H(\pi^\star)$. Enlarging the admitted query family adds task partitions to the common refinement, can only make $\pi^\star$ finer, and can never reduce $R_{\min}$. Proof: if two source states share an encoder value, exact decoding requires them to have the same answer for every query; this is precisely the refinement condition. Conversely, the condition defines a decoder on each encoder block. The $\pi^\star$ block identifier is a function of every sufficient $Z$, which gives the entropy bound, and adding queries can only add distinctions.
\end{proposition}

This proposition makes ``the codebook was designed for the wrong task'' precise. A representation may be sufficient for a narrow family and insufficient after the family expands even if its bit length does not change. Bit count alone is therefore not a fidelity guarantee: the induced distinctions must align with the declared query family.

\subsection{Linear-Logarithmic Write-Time and Read-Time Separation}

Consider $X=(X_1,\ldots,X_k)$ with independent uniform bits, let $Q$ identify one of the $k$ coordinates, and let $Y=X_Q$. A write-time encoder does not observe the realized coordinate.

\begin{proposition}[Linear-logarithmic separation under source fallback]
Any deterministic write-time representation that answers every coordinate query exactly satisfies $H(Z)\geq k$ bits. If the original versioned source remains addressable at read time, an exact path can instead send a coordinate address using $ceil(log_2 k)$ bits and return one evidence bit, for $R_{\mathrm{roundtrip}}=ceil(log_2 k)+1$. Hence
$R_{\mathrm{write}}/R_{\mathrm{roundtrip}}\geq k/(1+ceil(log_2 k))=Theta(k/log_2 k)$.
Proof: any two distinct bit vectors differ in some coordinate, so a representation sufficient for all coordinate queries must distinguish all source vectors and has entropy at least $H(X)=k$. The read-time path encodes the requested coordinate and returns its source bit. The comparison is between persistent query-agnostic representation and one online request-and-return exchange; it does not count the raw source store, index maintenance, authorization checks, or access latency as zero.
\end{proposition}

The result is linear-logarithmic in the number of fields $k$, not exponential in $k$. It becomes exponential only if the address length $n=log_2 k$ is chosen as the independent parameter, in which case the write-time lower bound is $2^n$ and the online exchange is $n+1$. The architectural implication is limited but useful: preserving a governed source-fallback path can avoid forcing one derivative representation to anticipate the joint distinctions of every future query.

\subsection{Indistinguishability, Addressability, Order, and Slicing}

\begin{proposition}[Interface indistinguishability]
Any storage or retrieval interface induces an equivalence relation over source states: two states are equivalent when every accessible result presented to the decoder is identical. If equivalent states have different answers for some admitted query, no downstream decoder restricted to that interface can be exactly sufficient. Proof: the decoder receives the same input for both states and therefore must return the same output for both.
\end{proposition}

This elementary proposition covers several discrete failure modes without claiming that they occur in every implementation. A lexical, positional, relational, or embedding-based index is adequate only to the extent that its complete accessible state preserves the distinctions required by the task family. A coarse retrieval cell that merges task-distinguishable values is insufficient; an embedding system that retains stable source identifiers and supports exact fallback need not have that defect. Addressability is thus a property of the whole interface and fallback path, not a label attached to one algorithm.

Order loss is the same phenomenon under a group action. If $Z(x)=Z(gx)$ for a permutation $g$ but an admitted task assigns different answers to $x$ and $gx$, the representation is insufficient. For $n$ distinct items under a uniform unknown ordering, a permutation-invariant representation discards $log_2(n!)$ bits of order identity. This lower bound applies only when order is task-relevant; invariant tasks legitimately quotient it away.

Slicing can be represented as an evidence hypergraph. Atomic source spans are vertices and each query has one or more sufficient evidence sets. Under a retrieval budget of at most $b$ slices, a fixed partition fails for a query whenever every sufficient evidence set intersects more than $b$ independently retrievable slices or contains a relation the interface does not expose. Larger chunks can change that incidence pattern but do not guarantee repair; the relevant object is the query-evidence structure, not chunk size alone.

\subsection{Architectural Consequences and Falsification Boundary}

The formal results support five design constraints for the proposed substrate. First, every derivative must retain resolvable source and snapshot identity. Second, the system must declare the task family for which a representation or policy is claimed sufficient. Third, indexes and summaries should be evaluated by the distinctions and evidence paths they preserve, not by compression ratio alone. Fourth, fallback requires executable access, authorization, freshness, provenance, and version compatibility in addition to retained bytes. Fifth, a widened task family or changed source distribution triggers revalidation because the minimal sufficient partition may have changed.

What remains empirical is at least as important. The appendix does not show that the Data Wiki reduces latency or cost, that its indexes outperform direct reading, that embedding retrieval is generally inferior, or that source fallback is operationally reliable. Those questions belong to P16's held-out comparison and to lifecycle-cost measurement. The proofs establish only a boundary: under the stated finite deterministic assumptions, once an interface merges two source states that an admitted task must distinguish, downstream intelligence alone cannot restore exactness.

\bibliographystyle{plain}
\bibliography{references}

\end{document}